\documentclass{bmvc2k}

\usepackage{graphicx}
\usepackage{tikz}
\usepackage{comment}
\usepackage{amsmath,amssymb} %
\usepackage{color}
\usepackage{float}

\usepackage{adjustbox}
\usepackage{stackengine}
\usepackage{multirow}

\usepackage{pifont}
\usepackage{booktabs}
\usepackage{tabularx}
\usepackage{wrapfig}
\usepackage{multirow}
\usepackage[capitalize]{cleveref}
\crefname{section}{Sec.}{Secs.}
\Crefname{section}{Section}{Sections}
\Crefname{table}{Table}{Tables}
\crefname{table}{Tab.}{Tabs.}
\usepackage{xspace}
\usepackage{csquotes}

\usepackage[hang,flushmargin]{footmisc}
\usepackage{enumitem}  %
\usepackage{colortbl}
\usepackage{xcolor}

\usepackage{tikz}
\usetikzlibrary{shapes,backgrounds}

\newcommand*{\myparagraph}[1]{\noindent\textbf{#1}\hspace{0.5em}}

\newcommand{\etc}{\emph{etc.}\@\xspace}
\newcommand{\eg}{\emph{e.\thinspace{}g.}\@\xspace}
\newcommand{\ie}{\emph{i.\thinspace{}e.}\@\xspace}

\newcommand{\etal}{\emph{et al.}\@\xspace}

\newcommand{\prob}{\ensuremath{\mathcal{P}}}
\newcommand{\papername}{$S^2$-Flow}
\newcommand{\md}{\ensuremath{\mathcal{D}}}
\newcommand{\mx}{\ensuremath{I}}

\newcommand{\xisty}{\ensuremath{I_{\text{sty}}}}
\newcommand{\xism}{\ensuremath{I_{\text{struct}}}}
\newcommand{\wplus}{\ensuremath{w^{+}}}
\newcommand{\wiplus}{\ensuremath{w}}

\newcommand{\wplust}[1]{\ensuremath{w_#1^{+}}}
\newcommand{\z}[1]{\ensuremath{z_{#1}}}
\newcommand{\mask}[1]{\ensuremath{M_{#1}}}

\newcommand{\ow}{\ensuremath{w}}
\newcommand{\ew}{\ensuremath{\hat{w}}}

\newcommand{\omask}{\ensuremath{m}}
\newcommand{\emask}{\ensuremath{\hat{m}}}

\newcommand{\oimg}{\ensuremath{I}}
\newcommand{\eimg}{\ensuremath{\hat{I}}}

\newcommand{\osm}{\ensuremath{w_{\text{sm}}}}
\newcommand{\esm}{\ensuremath{\hat{w}_{\text{sm}}}}

\newcommand{\osty}{\ensuremath{w_{\text{sty}}}}

\newcommand{\rot}[2][l]{\rotatebox[origin=#1]{90}{#2}} 
\definecolor{checkyes}{rgb}{1.0, 1.0, 1.0}
\definecolor{crossno}{rgb}{1.0, 1.0, 1.0}
\definecolor{maybe}{rgb}{1.0, 1.0, 1.0}
\def \y {$\checkmark$\cellcolor{checkyes}}
\def \n {$\times$\cellcolor{crossno}}
\def \l {$\circ$\cellcolor{maybe}}

\definecolor{turquoise}{cmyk}{0.65,0,0.1,0.3}
\definecolor{purple}{rgb}{0.65,0,0.65}
\definecolor{dark_green}{rgb}{0, 0.5, 0}
\definecolor{orange}{rgb}{0.8, 0.6, 0.2}
\definecolor{dark_orange}{rgb}{0.7, 0.6, 0.3}
\definecolor{red}{rgb}{0.8, 0.2, 0.2}
\definecolor{darkred}{rgb}{0.6, 0.1, 0.05}
\definecolor{blueish}{rgb}{0.0, 0.3, .6}
\definecolor{light_gray}{rgb}{0.7, 0.7, .7}
\definecolor{pink}{rgb}{1, 0, 1}
\definecolor{cyan}{rgb}{0., 1, 1}

\definecolor{tabbestcolor}{rgb}{0.785, 0.851, 0.969}
\def \best {\cellcolor{tabbestcolor!85}}
\def \sbest {\cellcolor{tabbestcolor!30}}

\title{$S^2$-Flow: Joint Semantic and Style Editing of Facial Images} %

\usepackage{xr}
\addauthor{Krishnakant Singh }{krishnakant.singh@visinf.tu-darmstadt.de}{1}
\addauthor{Simone Schaub-Meyer}{simone.schaub@visinf.tu-darmstadt.de}{1, 2}
\addauthor{Stefan Roth}{stefan.roth@visinf.tu-darmstadt.de}{1, 2}

\addinstitution{
Department of Computer Science\\ 
TU Darmstadt}
\addinstitution{
hessian.AI
}

\runninghead{Singh \etal} {\papername{}: Joint semantic and style editing of facial images}

\begin{document}

\maketitle

 \vspace{-2em}
 \begin{figure}[!htb]
     \centering
     \includegraphics[height=0.35\textwidth]{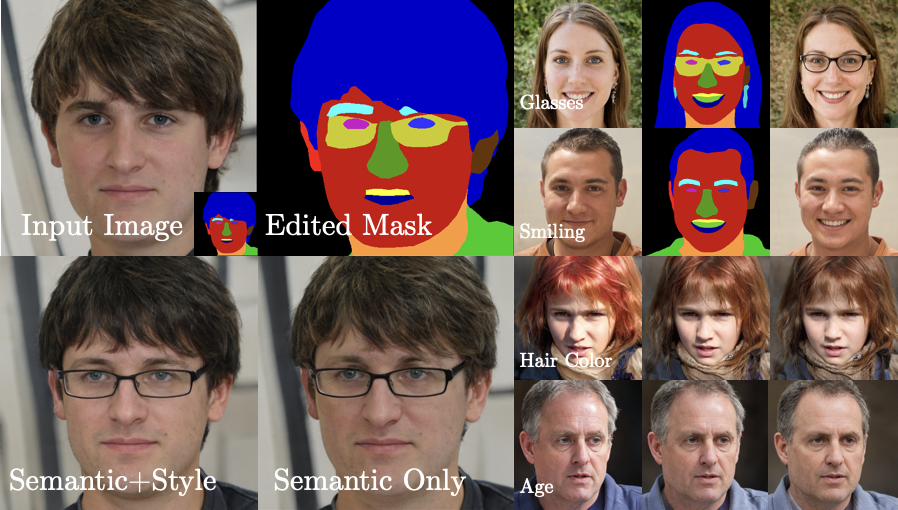}
     \vspace{-0.5em}
     \caption{\papername{} is capable of applying semantic \emph{(top right)}, style \emph{(bottom right)}, and joint semantic and style edits \emph{(left)} for facial images while preserving both identity and realism. 
     }
\label{fig:abstract:cover_image}
\vspace{-1em}
\end{figure}
\begin{abstract}

The high-quality images yielded by generative adversarial networks (GANs) have motivated investigations into their application for image editing. 
However, GANs are often limited in the control they provide for performing specific edits. 
One of the principal challenges is the entangled latent space of GANs, which is not directly suitable for performing independent and detailed edits. 
Recent editing methods allow for either controlled style edits \emph{or} controlled semantic edits. In addition, methods that use semantic masks to edit images have difficulty preserving the identity and are unable to perform controlled style edits. 

We propose a method 
to disentangle a GAN's latent space into semantic and style spaces, enabling controlled semantic \emph{and} style edits for face images independently within the same framework.  
To achieve this, we design 
an encoder-decoder based network architecture (\papername{}), which incorporates two proposed inductive biases.
We show the 
 suitability of \papername{} quantitatively and qualitatively by performing various semantic and style edits. Code and data are available at \href{https://github.com/visinf/s2-flow}{https://github.com/visinf/s2-flow}.

\end{abstract}
\renewcommand{\prob}{\ensuremath{\mathcal{P}}}
\renewcommand{\md}{\ensuremath{\mathcal{D}}}
\renewcommand{\mx}{\ensuremath{I}}

\renewcommand{\xisty}{\ensuremath{I_{\text{sty}}}}
\renewcommand{\xism}{\ensuremath{I_{\text{struct}}}}
\renewcommand{\wplus}{\ensuremath{w^{+}}}
\renewcommand{\wiplus}{\ensuremath{w_{I}}}

\renewcommand{\wplust}[1]{\ensuremath{w_#1^{+}}}
\renewcommand{\z}[1]{\ensuremath{z_{#1}}}
\renewcommand{\mask}[1]{\ensuremath{M_{#1}}}

\section{Introduction} \label{sec:intro}
Powerful deep generative models of images, such as generative adversarial networks (GANs) \citep{Goodfellow2014Generative} or variational autoencoders (VAEs) \citep{Kingma2014Auto}, have seen numerous applications across computer vision \citep{Peebles2022Gan,Mu2022Coordgan,Abdal2021Labels4Free,Abdal2022Video2StyleGAN,Zhang2021Datasetgan}. 
With the advent of models based on StyleGAN~\citep{Karras2019Style}, there has been a plethora of work focusing on controllable manipulation of the latent code for the task of image editing \citep{Collins2020Editing,Haerkoenen2020GANSpace,Shen2022InterFaceGAN}.
However, modifying the latent code in a controllable way, such that it leads to the desired edits in the image space, continues to be challenging.

Broadly speaking, we can divide image editing with GANs into two subgroups: 
\emph{(i) Unconditional GAN-based methods}~\citep{Wu2021Stylespace, Haerkoenen2020GANSpace}, which find editing vectors using unsupervised learning methods like PCA~\citep{Haerkoenen2020GANSpace} or activation maps~\citep{Wu2021Stylespace}. They do not take user inputs into account and have a limited set of editing directions.
\emph{(ii) Conditional GAN-based methods}, on the other hand, are more cognizant to the user input. These methods generate the edited image conditioned on user inputs, such as semantic masks~\citep{Lee2020MaskGAN,Park2019Semantic,Zhu2020SEAN}, attributes~\citep{He2019AttGAN,Hou2022GuidedStyle}, or text~\citep{Patashnik2021Styleclip,Xia2021Tedigan,Li2020Manigan}. 
Though these methods support more varied editing operations, they lack controllability, \eg, attributed-based methods provide no controllability on how a smile (wide, grinning, \etc) might look or if the person is wearing round or square glasses (\cref{fig:related_work:control_exp:left}). 
Semantic-based methods, in contrast, have limited control on style editing, requiring the use of a target transfer image (\cref{fig:related_work:control_exp:right}).
In general, performing controlled and disentangled edits in the latent space is a very challenging task.

In this paper, we propose, to the best of our knowledge, the first approach that allows to perform controlled semantic editing (\ie~changes possible with a semantic mask, \eg, changing smile, changing hair style, \etc) \emph{and} style editing (\ie~changes \emph{not} possible with a semantic mask, \eg, age, hair color, \etc) while preserving the identity of facial images.
We achieve this by disentangling the semantic and style spaces.
This disentanglement is achieved by introducing two inductive biases into the network:
\emph{(1) \textbf{Style consistency}} -- editing an image in the semantic domain should have no effect on the style properties of the image. 
\emph{(2) \textbf{Semantic consistency}} -- edits made in the semantic domain should be reflected in the semantics of the generated image. 
Our design of the model architecture and a novel loss formulation allow the model to incorporate the aforementioned inductive biases, helping it to make independent edits to the style and semantics of a given image. 
Specifically, our contributions are:
\emph{(i)} We propose a method to disentangle the latent space of a pretrained generator network into style and semantic spaces for the task of facial editing. Thereby, we are among the first to utilize normalizing flows for disentangling a GAN's latent space.
\emph{(ii)} Our method solves the problem of applying fine-grained edits to both the style \emph{and} semantic spaces.
\emph{(iii)} We show both qualitatively and quantitatively that our model outperforms well established methods~\citep{Lee2020MaskGAN,Park2019Semantic} on two semantic editing benchmarks.
\emph{(iv)} We show our model's ability to generate high-quality identity-preserving edits for various editing tasks, see \cref{fig:abstract:cover_image}.  

\section{Related Work}\label{sec:related_work}
Since the seminal work of Goodfellow \etal \cite{Goodfellow2014Generative}, there has been enormous progress in the area of generative modelling, from generating small-scale low-resolution images \cite{Radford2016Unsupervised} to generating high-fidelity, realistic looking images \cite{Karras2018Progressive,Karras2019Style}. 

In this work we tackle the problem of disentangling the latent space of a pretrained GAN, focusing on facial image editing.

\myparagraph{Disentanglement.} Unsupervised disentanglement has been a long standing goal for computer vision for its usefulness in inverting the generative process. 
InfoGAN~\citep{Chen2016Infogan} and BetaVAE~\citep{Higgins2017beta} tackle this problem from an information theoretic perspective.
Despite their pioneering efforts, these models work only in low-resolution low-complexity dataset settings and are difficult to train. 
Many works \citep{Huang2018Multimodal, Liu2019Few} disentangle the style and semantic codes of an image by swapping the codes with another image for the the task of image-to-image translation. 
Instead of swapping style and semantic codes, we obtain a disentanglement between these spaces by incorporating transformation-based inductive biases like semantic and style consistency. 
Recently, \citep{Xu2022Transeditor,Kwon2021Diagonal} proposed disentangled variants of StyleGAN~\citep{Karras2019Style} by using two separate (style and semantic) spaces and modelling the interactions between them using attention modules. 
In contrast, our work deals with disentangling the latent space of a pretrained GAN and, thus, requires no additional training of the GAN module, which is difficult and computationally expensive. 

\begin{wraptable}[14]{rt}{0.58\linewidth}
\centering
\scriptsize
\setlength{\tabcolsep}{1pt}
\def\arraystretch{0.8}
\vspace{-1.5em}
\begin{tabular}{lcccc|cccc|cccc|ccc|c}
\toprule
& \multicolumn{4}{c}{Attribute}
& \multicolumn{4}{c}{Semantic}
& \multicolumn{4}{c}{Uncondit.}
& \multicolumn{3}{c}{Text}
& Ours 
\\
\cmidrule(lr){2-5}
\cmidrule(lr){6-9}
\cmidrule(lr){10-13}
\cmidrule(lr){14-16}
\cmidrule(lr){17-17}
& 
\rot{\tiny{StyleFlow~\citep{Abdal2021StyleFlow}}}&
\rot{\tiny{InterfaceGAN~\citep{Shen2022InterFaceGAN}}}&
\rot{\tiny{LTNeuralODE~\citep{Khrulkov2021Latent}}}&
\rot{\tiny{AttGAN~\citep{He2019AttGAN}}}&
\rot{\tiny{SPADE~\citep{Park2019Semantic}}}&
\rot{\tiny{SEAN~\citep{Zhu2020SEAN}}}&
\rot{\tiny{MaskGAN~\citep{Lee2020MaskGAN}}}&
\rot{\tiny{EditGAN~\citep{Ling2021EditGAN}}}&
\rot{\tiny{Stylespace~\citep{Wu2021Stylespace}}}&
\rot{\tiny{GANSpace~\citep{Haerkoenen2020GANSpace}}}&
\rot{\tiny{Steerability~\citep{Jahanian2020Steerability}}}&
\rot{\tiny{GANInvers.~\citep{Abdal2019Image2stylegan,Abdal2020Image2stylegan++,Zhu2020Domain}}}&
\rot{\tiny{StyleClip~\citep{Patashnik2021Styleclip}}}&
\rot{\tiny{TediGAN~\citep{Xia2021Tedigan}}}&
\rot{\tiny{ManiGAN~\citep{Li2020Manigan}}}&
\rot{\tiny{\papername{}}} \\
\midrule
\scriptsize{Style Edits} & \y & \y & \y & \y & \l & \l &\l & \n & \l & \l & \l & \l & \y & \y & \y & \y\\
\scriptsize{Structural Edits} & \l & \l & \l & \l & \y & \y & \y & \y & \l & \l & \l & \l & \l & \l & \l & \y\\
\scriptsize{Mult.~Edits / Model} & \y & \n & \n & \y & \y & \y &\y & \l & -- & \n & \n & -- & \y & \y & \y & \y\\ 
\scriptsize{Pretrained GAN} & \y & \y & \y & \n & \n & \n & \n & \y & \y & \y & \y & \y & \y & \n & \n & \y\\
\bottomrule
\multicolumn{16}{l}{\tiny{$\checkmark$: Fully, $\times$: None, $\circ$: Limited (requires style image), $-$: N/A}} \\
\end{tabular}
\smallskip
\caption{Comparison of existing editing methods. Our work sits between attribute and semantic methods, allowing for both fully controlled style \emph{and} semantic edits.}
\label{tbl:related_work:comparision_tbl}
\end{wraptable}

\begin{figure}[t]
    \centering
    \subfigure[Controlled semantic editing]{
    \begin{minipage}[t][0.25\textwidth][t]{0.45\textwidth}
        \includegraphics[width=\textwidth]{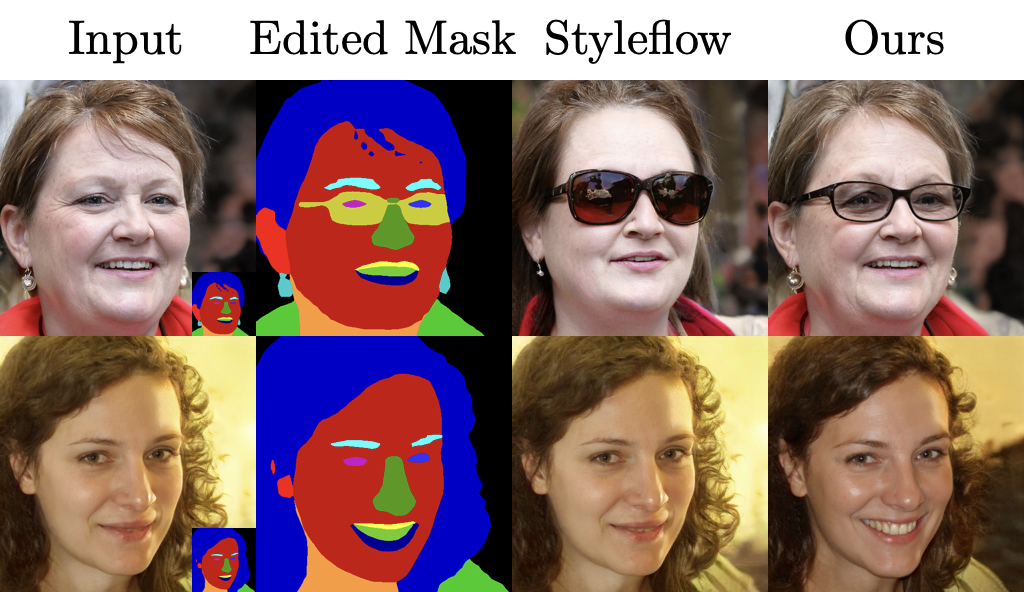}
    \end{minipage}
    \label{fig:related_work:control_exp:left} 
    }
    \hspace{.05\textwidth}
    \subfigure[Controlled style editing]{
    \begin{minipage}[t][0.25\textwidth][t]{0.45\textwidth}
        \includegraphics[width=\textwidth]{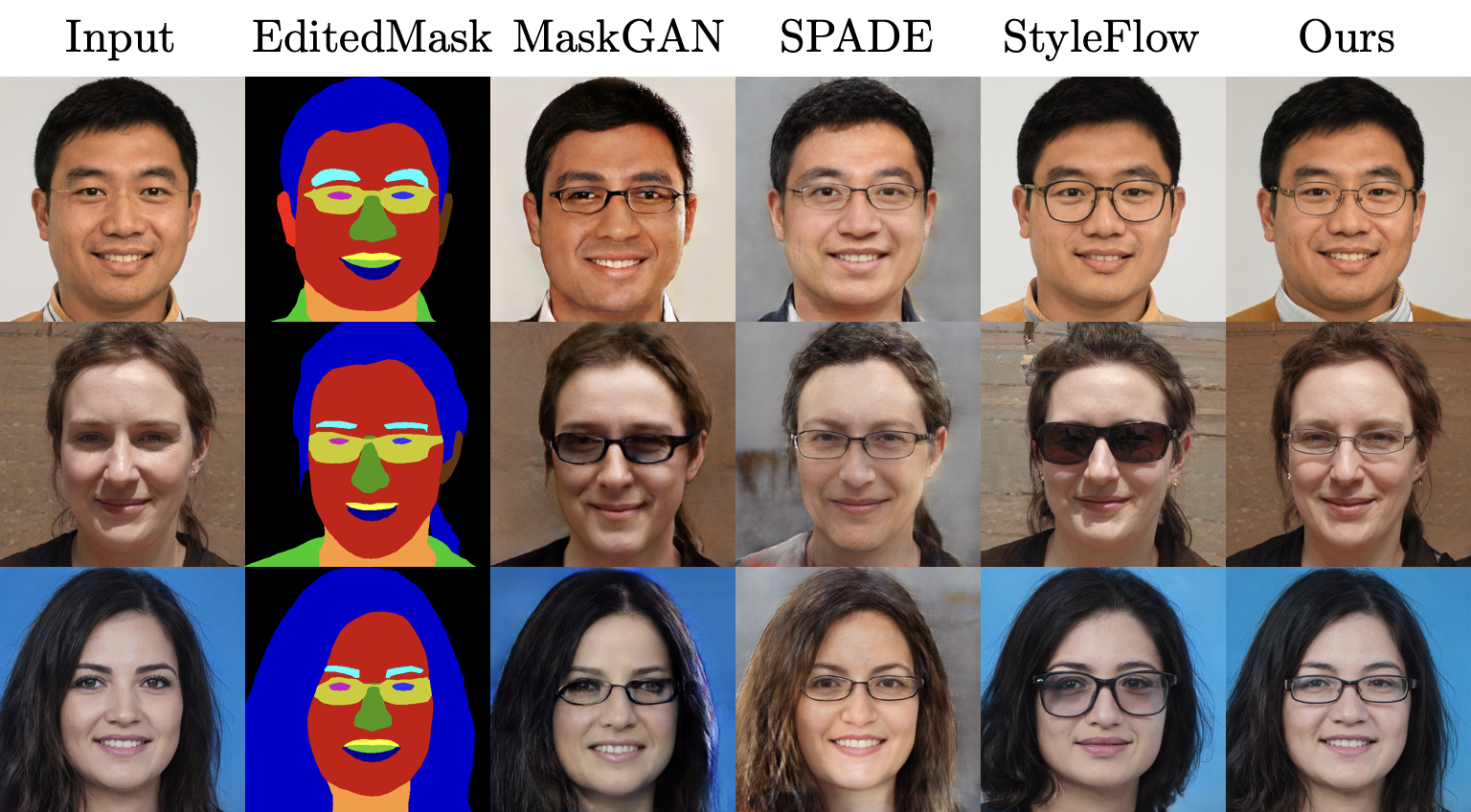}
        \label{fig:related_work:control_exp:right}
    \end{minipage}
    }
    \caption{\emph{(a)} \textbf{Controlled semantic editing.}  Attribute-based methods, \eg StyleFlow \citep{Abdal2021StyleFlow}, do not allow for user control over how a targeted semantic edit should look like.
    \emph{(b)} \textbf{Controlled style editing.} Semantic methods, \eg MaskGAN \citep{Lee2020MaskGAN}, allow for limited controllability for style editing.
    As semantic-based methods rely on target images for style editing, erroneous attributes like lip makeup/skin color are also transferred when the user only aimed to change the the age (\emph{top}) or hair color (\emph{bottom}). In contrast, \papername{} does not require target images and uses interpolation in style space, enabling it to apply targeted style edits.}
    \vspace{-0.5em}
    \label{fig:related_work:control_exp}
\end{figure}

\myparagraph{Editing in the latent space of GANs.} 
There have been a plethora of interesting works for editing using the latent space of GANs. 
Attribute-based methods rely on learning interpretable edit directions in the latent space of a pretrained GAN using a neural network trained either on attributes \citep{Shen2022InterFaceGAN,Abdal2021StyleFlow,Khrulkov2021Latent,Zhuang2021Enjoy}, paired synthetic data \citep{Viazovetskyi2020StyleGAN2}, or pseudo labels \citep{Jahanian2020Steerability}. 
Some attribute-based methods like AttGAN~\cite{He2019AttGAN} train a generative model based on an attribute classification loss. 
Text-based methods use natural language cues for editing the image; \citet{Patashnik2021Styleclip} apply a CLIP-based \citep{Radford2021Learning} loss for learning editing directions. 
\citep{Xia2021Tedigan,Li2020Manigan} learn a GAN-based model conditioned on text. 
All the above methods provide a high degree of control for style edits but severely lack controllability and interpretability in terms of editing semantics, \eg, sunglasses \emph{vs.}\ reading glasses, small smile \emph{vs.}\ big smile, \etc (\cref{fig:related_work:control_exp:left}).
A fastidious user can give very detailed textual cues to obtain the desired effect, but this soon becomes untenable when finer control of semantics is required, \eg, when describing the exact shape of glasses.
On the other hand, methods like \cite{Collins2020Editing,Haerkoenen2020GANSpace,Shen2021Closed,Voynov2020Unsupervised,Wu2021Stylespace} use unsupervised approaches for finding editing directions. 
These methods find a set of editing directions and require copious manual effort to semantically identify them. 
Semantic-based editing methods \citep{Lee2020MaskGAN,Park2019Semantic, Zhu2020SEAN} provide control over semantic editing but these models severely lack in the ability to carry out style edits. For carrying out targeted style edits, the user has to search through 1000s of images to find a suitable target (\cref{fig:related_work:control_exp:right}).
An overview of such work is shown in \cref{tbl:related_work:comparision_tbl}.

\myparagraph{Simultaneous edits of semantic and style.}
Our work is able to provide controlled semantic and style editing on facial images by finding disentangled dual spaces.
Unlike \citep{Xu2022Transeditor,Kwon2021Diagonal,Kazemi2019Style}, we work on a pretrained GAN's latent space instead of training a new GAN model from scratch, which requires a large training set and extensive parameter tuning. 
Our method of disentanglement uses only 10k GAN-generated images. 
Moreover, it lies at the intersection of style and semantic-based methods, using semantic conditioning for enabling semantic edits and learning walks in the style space for applying highly controllable style edits.

\section{Joint Semantic and Style Editing}\label{sec:method}
Given \emph{only} a pretrained GAN model and its generated images as input, our goal is to devise a method for image editing that enables semantic and style edits within the \emph{same} framework without mutual interference between these edits.
We argue that this requires disentangling the image representation into two parts, one responsible for the semantics of the image and one for capturing its style.
This disentanglement allows us, on one hand, to perform edits in multiple spaces, namely in the semantic and style space, while on the other giving more control on the generated result by ensuring the integrity of the non-edited characteristics.

The key insight of our method is that an image can be decomposed into its semantic and the style 
codes and edits made in one domain (semantic or style) should not affect the other. 

We design our image generation and editing framework as an encoder-decoder model to disentangle the aforementioned two spaces 

using continuous normalizing flow (CNF)~\cite{Chen2018Neural} blocks. We chose CNFs based on two motivations:
\emph{(i)} Firstly, a CNF network is reversible by design and hence has cycle consistency, which is a crucial property to successfully disentangle semantics and style during training.
\emph{(ii)} Secondly, CNFs are much easier to train than other encoder-decoder models like VAEs \cite{Higgins2017beta,Kingma2014Auto} or Transformers~\cite{Vaswani2017Attention}.
Before describing our architecture in \cref{sec:model}, we summarize its building blocks.

\subsection{Building blocks}
\label{sec:background}
\myparagraph{Latent space.}
The latent space of deep generative models serves as a good proxy for the real image manifold.
We use the latent space of StyleGAN2~\cite{Karras2019Style} for our model. 

Given a latent sample, drawn from $\mathcal{N}(0, I)$, StyleGAN2 transforms it into an intermediate latent code 
using a series of nonlinear mappings. \citet{Abdal2019Image2stylegan} further extend this space by concatenating 18 different latent codes, which they term the $\mathcal{W}^{+}$ space. We train our network directly in this $\mathcal{W}^{+}$ space, since \cite{Abdal2020Image2stylegan++,Abdal2021StyleFlow} show that this space is better suited for editing.

\myparagraph{Normalizing flows.}
A normalizing flow model \cite{Dinh2017Density} transforms a simple initial known distribution to a more complex one using a series of
composable transformations $\mathit{f_1}, \ldots ,\mathit{f_k}$. 
The function $F=\mathit{f_1}\circ \cdots \circ \mathit{f_k}$ must be invertible and both $F$ and $F^{-1}$ should be differentiable. 
The function $F$ relates the marginal densities of the two distributions using the change of variables formula, which involves computing the determinant of the Jacobian.
Calculating the determinant of the Jacobian can be an expensive operation, requiring special neural network architectures for fast computation. 
\citet{Chen2018Neural} introduced a continuous version to alleviate this problem, paving the way for using arbitrary neural networks for modelling these transformations. 

We decided to use CNFs~\cite{Chen2018Neural} for our implementation as this allows us to use an unrestricted architecture when modelling the transformation function $F$, making the transformation function more flexible and expressive. 
\subsection{\papername{} model}
\label{sec:model}

\begin{figure}[t]
    \centering
    \subfigure[Training]{
    \begin{minipage}[t][0.25\textwidth][t]{0.70\textwidth}
        \includegraphics[width=\textwidth]{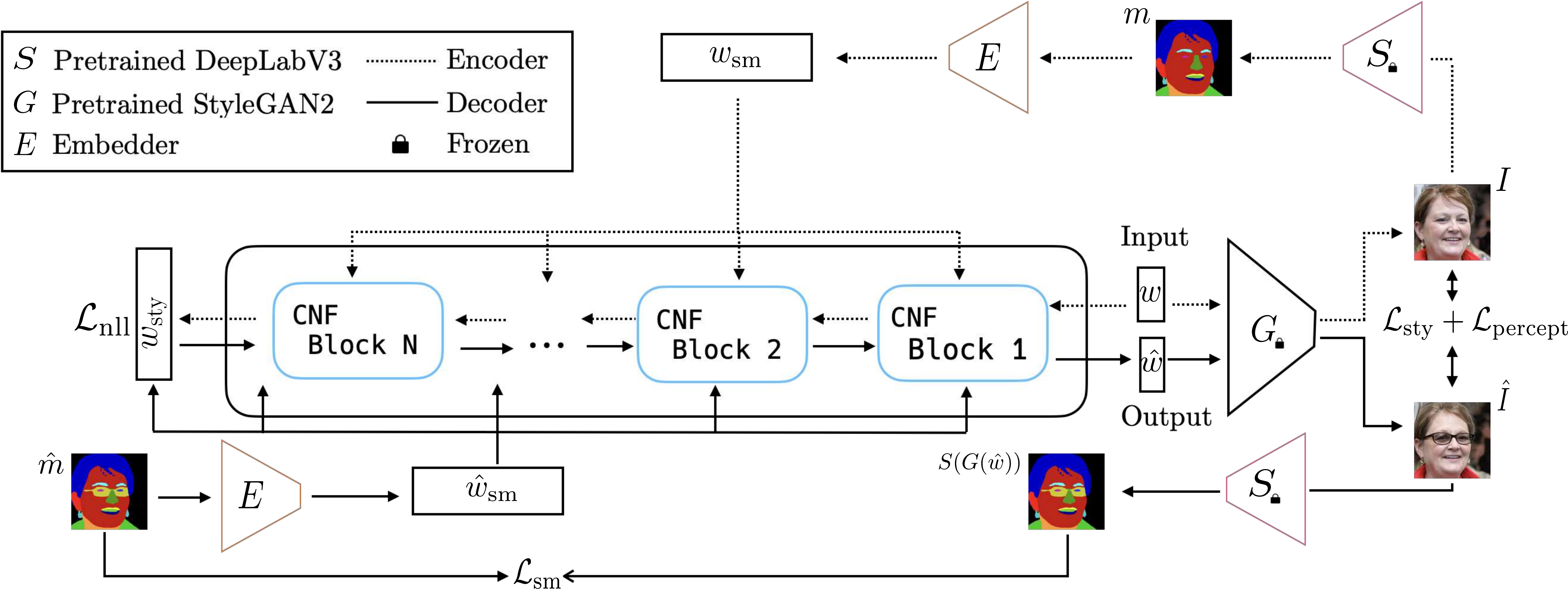}
    \end{minipage}
    \label{fig:method:network_arch}
    }
    \hspace{5pt}
    \subfigure[Editing a real image]{
    \begin{minipage}[t][0.25\textwidth][t]{0.20\textwidth}
        \includegraphics[width=\textwidth]{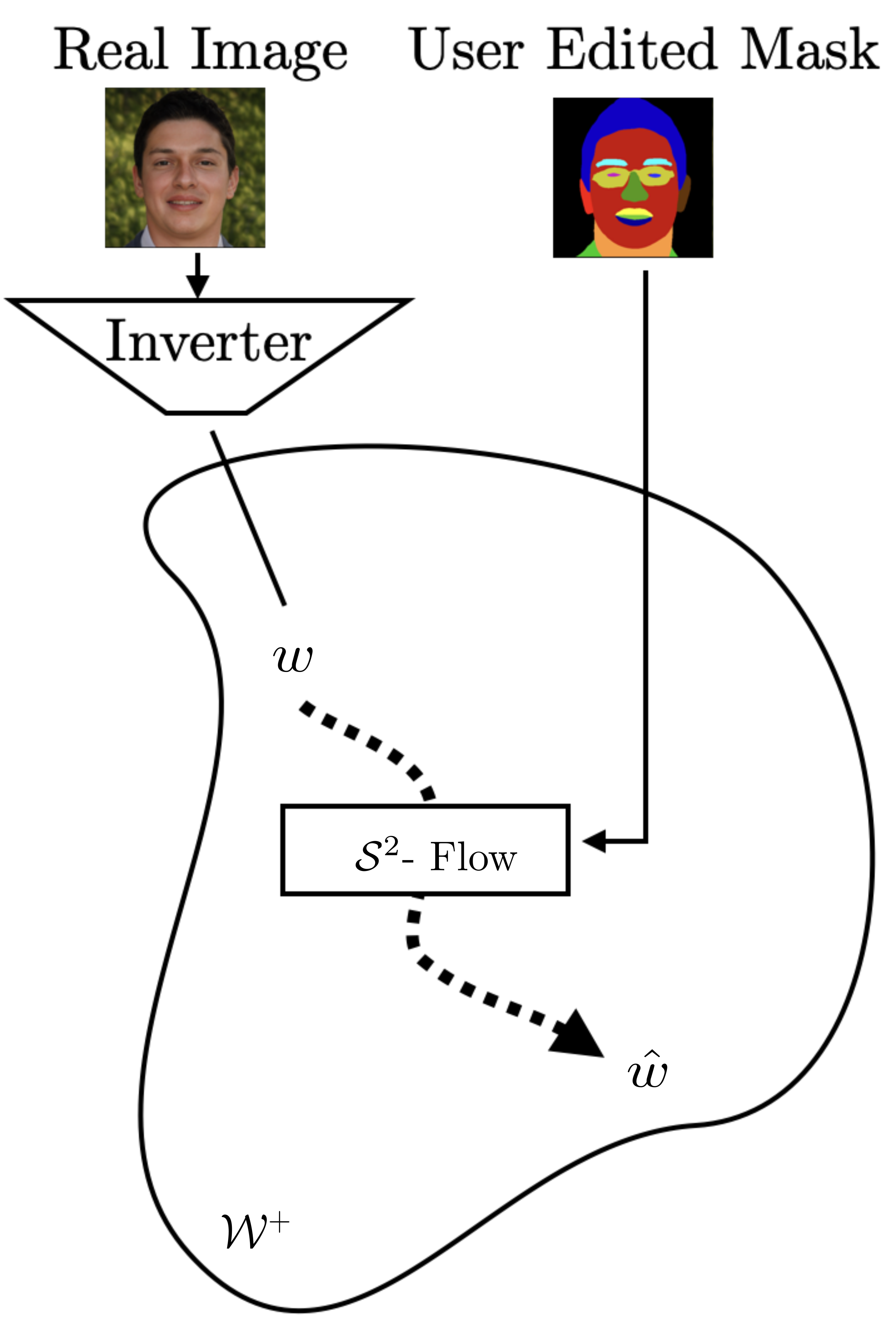}
        \label{fig:method:latent_walk}
    \end{minipage}
    }
    \caption{\textbf{Framework overview.} \emph{(a)} \textbf{Training.} (1) The encoder \emph{(dashed line)} takes as input the latent code \ow\ $\in \mathcal{W}^{+}$ and its corresponding inferred semantic mask \omask\ = $S(G(\ow))$. (2) The encoder model disentangles the style code \osty\ from \ow\ given \osm = $E(m)$. (3) The decoder (\emph{solid line}) combines \osty\ and the edited semantic code \esm = $E$(\emask) to yield the edited latent code \ew\ $\in \mathcal{W}^{+}$ (output), which is fed to the generator network to yield the edited image \eimg. \emph{(b)} \textbf{Editing.}  Given a real image, we first obtain its latent code \ow{} using an inverter network, \eg, e4e~\citep{Tov2021Designing}. Using this \ow\ and the given user-edited mask, \papername{} returns a new latent code \ew, which is used to generate the edited image $G(\ew)$.
    }
    \vspace{-0.5em}
\end{figure}
\myparagraph{Overview.}
The goal of \papername{} is to disentangle the latent code \ow\ into its constituent latent codes, namely, semantic \osm\ and style \osty. 
We use a conditional variant of CNF~\cite{Chen2018Neural}, where the forward and reverse flow\footnote{CNFs define forward flow as transforming a normal distribution to a more complex one. The reverse flow transforms a complex distribution to the normal distribution.} are used to model the decoder and encoder, respectively. 
In contrast to earlier works \cite{Abdal2021StyleFlow, Khrulkov2021Latent}, where the latent space of the CNF model is highly entangled, our formulation disentangles style and semantics using our novel inductive biases (\cref{sec:objective}). 
The process of encoding-decoding is visualized in \cref{fig:method:network_arch} and formally described below.

\myparagraph{Encoder.} 

The reverse flow of our model is the encoder network and works as follows: 
Given a latent code \ow, we generate its corresponding image \oimg\ and infer its semantic mask \omask\ using pretrained StyleGAN2~\citep{Karras2019Style} and DeepLabV3~\citep{Chen2018Encoder} networks, respectively. 
The semantic mask is then passed through a convolutional neural network called Embedder to yield the semantic code \osm. 
The concatenation of  \ow\ and \osm\ is fed as input to each CNF block of our model. 
Given this semantic code \osm, the encoder learns to disentangle the style code \osty\ from the latent code \ow.

\myparagraph{Decoder.} 
Our forward flow learns to combine the latent style (\osty) and semantic (\osm) codes to reconstruct the original latent code \ow\

without any loss of information due to the reversibility of the CNF.

During training (\cref{fig:method:network_arch}), we simulate the editing behaviour by modifying the input mask \omask\ to \emask, which results in a new \esm, leading to an edited latent code \ew\ and its corresponding edited image \eimg=$G(\ew)$.
Editing during training is performed by swapping the mask between two samples based on our defined criterion, see supplemental for more details.
\subsection{Objective}
\label{sec:objective}

To disentangle style from semantics for the $\mathcal{W}^{+}$ space during training, we add two inductive biases to our loss function: 
\emph{(i) Style consistency} ensures 
that editing the semantic mask of an image should have minimal impact on the style attributes. 
Hence at training time, the decoder should output a latent code \ew\ that is similar in style to the original \ow. 
In other words, \papername{} learns to extract the same style code \osty\ for two images that only differ semantically. 
\emph{(ii) Semantic consistency} 
addresses the fact that changes made in the semantic domain should be reflected in the generated image as well. 
Our overall loss function is defined as
\begin{equation}
      \mathcal{L} = \mathcal{L}_\text{nll}(\ow) + \lambda_{1} \mathcal{L}_\text{sm}(\emask, S(G(\ew))) + \lambda_{2} \mathcal{L}_\text{img}(\oimg, \eimg) + \lambda_{3} \mathcal{L}_\text{percept}(\oimg, \eimg) \ .
    \label{eqn:full_loss}
\end{equation}
The negative log-likelihood loss, $\mathcal{L}_\text{nll}$, encourages the model to learn the conditional data distribution of images given semantic masks.  

To ensure semantic consistency, we use the $\mathcal{L_\text{sm}}$ loss, which is equal to the cross-entropy loss between the edited mask $\emask$ and the inferred mask of the generated image $S(G(\ew))$. 
To ensure style consistency between the edited image \eimg\ and the original image \oimg, we use two loss functions,  $\mathcal{L}_\text{img}$ and $\mathcal{L}_\text{percept}$. $\mathcal{L}_\text{img}$ measures the $L_2$ distance in the image space. We use a masked version of the $L_2$ distance, restricting the computation of the loss only to the edited regions.

The perceptual loss, $\mathcal{L}_\text{percept}$, computes the $L_1$ distance between \eimg\ and \oimg\ using the intermediate features from a pretrained VGG network \citep{Simonyan2015Very}. The detailed formulas can be found in the supplemental.

\subsection{Editing and generation} 
After training, \papername{} can be used for both conditional image generation and editing.

\myparagraph{Conditional generation.}
Given a semantic mask \omask{} and a style code \osty\ from the style space of \papername{}, the decoder generates \ow\ and consequently a new image \oimg=$G(\ow)$ consistent with the semantic mask \omask.

\myparagraph{Editing.}
Given an edited mask for an image \oimg\ (real or fake) and the latent code\footnote{For a real image, we use e4e~\citep{Tov2021Designing} to obtain the corresponding latent code.} \ow, the encoder disentangles \ow\ into \osty\ and \osm.
For semantic editing, the decoder uses the edited mask and the original style code \osty\ to create the edited latent code in $\mathcal{W}^{+}$. \cref{fig:method:latent_walk} shows an illustrative example of editing a real image. 

Style editing is performed by linearly interpolating between the source style code \osty\ and the target style code in the style latent space. The target style code equals the mean style code of all positive samples for the given target attribute. 
Semantic and style edits can also be applied simultaneously, see \cref{fig:abstract:cover_image}.

\section{Experiments}
To evaluate the disentangled editing ability of our approach, we perform various facial editing experiments by applying style and semantic edits separately and in combination. We compare the results visually and quantitatively to related editing methods \citep{Abdal2021StyleFlow,Lee2020MaskGAN,Park2019Semantic}.

\myparagraph{Dataset and training.} The main goal of our work is to disentangle the latent space of a pretrained GAN by training our model on GAN-generated images. In particular, we use the dataset introduced by StyleFlow~\citep{Abdal2021StyleFlow}, which consists of 10k latent codes from a StyleGAN2~\citep{Karras2019Style} model trained on FFHQ~\citep{Karras2019Style}.

For training we use the Adam~\citep{Kingma2015Adam} optimizer with a constant learning rate of $3\cdot 10^{-4}$. Further, we rely on a curriculum learning approach where the loss function and the difficulty of the performed edits in the segmentation mask are gradually increased. We refer the reader to the supplementary for more training details. 

\myparagraph{Metrics.} 
For measuring the structural similarity between the edited semantic mask and the semantic mask of the generated image, we use the mean IoU (mIoU) and the mean pixel-wise accuracy (mAcc). 
To compare the quality of generated images between different models, we use FID~\citep{Heusel2017GANs} and LPIPS~\citep{Zhang2018Unreasonable}. 

We use the ArcFace~\citep{Deng2019ArcFace} network to measure the identity preservation score (ID) when editing an image.

\myparagraph{GAN inversion.} 
For editing real images, our model makes use of GAN inversion methods \citep{Tov2021Designing,Abdal2019Image2stylegan,Abdal2020Image2stylegan++,Richardson2021Encoding,Zhu2020Domain}; specifically, we use \citep{Tov2021Designing} to obtain the latent code of a real image.
\begin{table}
\parbox[t][][t]{.49\linewidth}{
    \def\arraystretch{0.2}
    \setlength{\tabcolsep}{1pt}
    \scriptsize
    \begin{tabularx}{1\linewidth}{@{}l*5{>{\centering\arraybackslash}X}@{}}
        \toprule
        & \multicolumn{2}{c}{Perceptual Quality} & \multicolumn{2}{c}{Semantic}&Identity\\
        \cmidrule(l{4pt}r{4pt}){2-3} \cmidrule(l{4pt}r{4pt}){4-5} \cmidrule(l{4pt}r{4pt}){6-6}    
        Method & FID $\downarrow$ & LPIPS $\downarrow$  & mIoU $\uparrow$ & mAcc $\uparrow$ &  ID  $\downarrow$\\     
        \midrule
            MaskGAN~\citep{Lee2020MaskGAN}  & 40.58 & 0.27 & 0.52 & 0.88 & 0.53  \\
            SPADE~\citep{Park2019Semantic}& 60.43 & 0.28 & 0.81 & 0.97 & 0.46 \\\midrule
            \papername{} \emph{(ours)} & 26.65 &  0.14 & 0.77 & 0.95 & 0.15 \\
            Abs. improv. & +13.93 &  +0.13 & -0.04 & -0.02 & +0.31 \\
        \bottomrule
        \end{tabularx}
\vspace{-1em}
\caption{Results on the smile edit benchmark from~\citep{Lee2020MaskGAN}}
\label{tbl:exp:smile_edit_benchmark}
}
\hfill
\parbox[t][][t]{.49\linewidth}{
\centering
    \scriptsize
    \def\arraystretch{0.2}
    \setlength{\tabcolsep}{1pt}
    \begin{tabularx}{1\linewidth}{@{}l*5{>{\centering\arraybackslash}X}@{}}
        \toprule
        & \multicolumn{2}{c}{Perceptual Quality} & \multicolumn{2}{c}{Semantic}&Identity\\
        \cmidrule(l{4pt}r{4pt}){2-3} \cmidrule(l{4pt}r{4pt}){4-5} \cmidrule(l{4pt}r{4pt}){6-6}    
        Method & FID $\downarrow$ & LPIPS $\downarrow$  & mIoU $\uparrow$ & mAcc $\uparrow$ &  ID  $\downarrow$\\
        \midrule
            MaskGAN~\citep{Lee2020MaskGAN} & 40.79 & 0.29 & 0.58 & 0.82& 0.50\\
            SPADE~\citep{Park2019Semantic}& 61.02 & 0.30 & 0.89 & 0.92 & 0.46\\\midrule
            \papername{} \emph{(ours)} & 26.75 & 0.12 &  0.78 & 0.94 & 0.10\\
             Abs. improv. & +14.04 & +0.17 &  -0.11 & +0.02 & +0.36\\            
        \bottomrule
        \end{tabularx}
\vspace{-1em}
\caption{Results on our general editing benchmark}
\label{tbl:exp:general_benchmark}
}
\vspace{-0.5em}
\end{table}

\begin{figure}[t]
\centering
\subfigure[Glasses editing]{
\begin{minipage}[t][0.20\textwidth][t]{0.40\textwidth}
\centering
\includegraphics[width=\textwidth]{results/controlled_editing/conent_control_exp.png}
\end{minipage}
\label{fig:exp:sematic_editing_qualatative:left}
}
\hspace{.01\textwidth}
\subfigure[Smile editing]{
\begin{minipage}[t][0.20\textwidth][t]{0.40\textwidth}
\centering
\includegraphics[width=\textwidth]{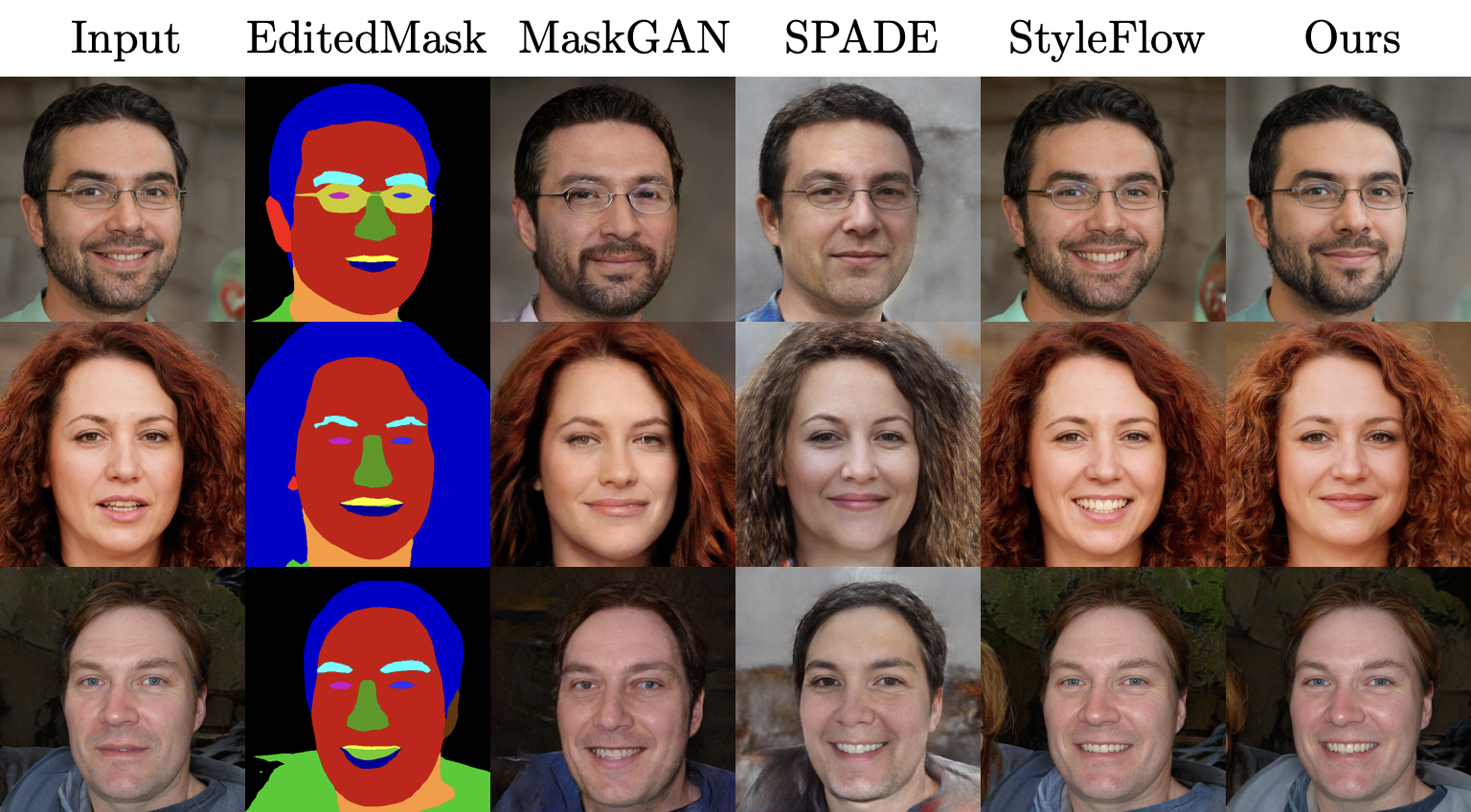}
\end{minipage}
\label{fig:exp:sematic_editing_qualatative:right}
}
\caption{\textbf{Comparison on semantic editing.} \papername{} successfully supports semantic edits while preserving the identity and yields higher quality images compared to existing semantic editing methods (MaskGAN \citep{Lee2020MaskGAN}, SPADE \citep{Park2019Semantic}). Also, \papername{} allows for more controlled user edits compared to existing attribute-based editing method (StyleFlow \citep{Abdal2021StyleFlow}).
}
\label{fig:exp:semantic_editing_qualatative}
\vspace{-0.5em}
\end{figure}

\subsection{Quantitative experiments}
\papername{} is a semantic-based method, which additionally allows for style editing, see \cref{tbl:related_work:comparision_tbl}. 
Thus, we compare our method against other semantic editing methods, namely MaskGAN \citep{Lee2020MaskGAN} and SPADE~\citep{Park2019Semantic}, for the task of semantic editing. Both these models perform highly controllable edits conditioned on semantic masks. 
We do not compare our method against EditGAN~\citep{Ling2021EditGAN} as it requires additional test-time optimization for each image for each editing direction; therefore, the same editing vector is not applicable to multiple images.
To quantitatively measure the editing capability, we use the smile edit benchmark introduced by MaskGAN~\citep{Lee2020MaskGAN}. 
For fair comparison, we train SPADE~\citep{Park2019Semantic} also on the StyleFlow~\citep{Abdal2021StyleFlow} dataset using the official repository. For MaskGAN~\citep{Lee2020MaskGAN}, we use the pretrained weights from the network trained on the CelebA-HQ dataset~\citep{Karras2018Progressive} provided by the authors. Both the CelebA-HQ and the StyleFlow dataset contain high-quality, diverse face images; hence, the test time distribution shift should be minimal.  
We report the results in \cref{tbl:exp:smile_edit_benchmark}. Our model outperforms both semantic editing methods in terms of ID, LPIPS, and FID while being minimally worse in terms of mIoU and mAcc scores. 
We also evaluate on a more complex editing scenario in which we make diverse edits to the semantic masks.
For each test image we randomly perform one of the following edits: \emph{(1)} swap mouth, \emph{(2)} swap nose, \emph{(3)} swap eyebrows, \emph{(4)} remove glasses, \emph{(5)} swap/add glasses, and \emph{(6)} swap hair. 
We refer to this as the general semantic editing benchmark; the results are shown in \cref{tbl:exp:general_benchmark}. 
\papername{} again outperforms all other baselines in terms of ID, LPIPS, and FID. 
Our model's slightly inferior mIoU and mAcc scores can be attributed to its behaviour of making more realistic and conservative edits to preserve the identity rather than only optimizing for the mIoU score (\cref{fig:exp:semantic_editing_qualatative}).
\subsection{Qualitative results}

\begin{figure}[t]
\centering
\subfigure[Fixed style]{
\begin{minipage}[t][0.20\textwidth][t]{0.43\textwidth}
\centering
\includegraphics[width=\textwidth]{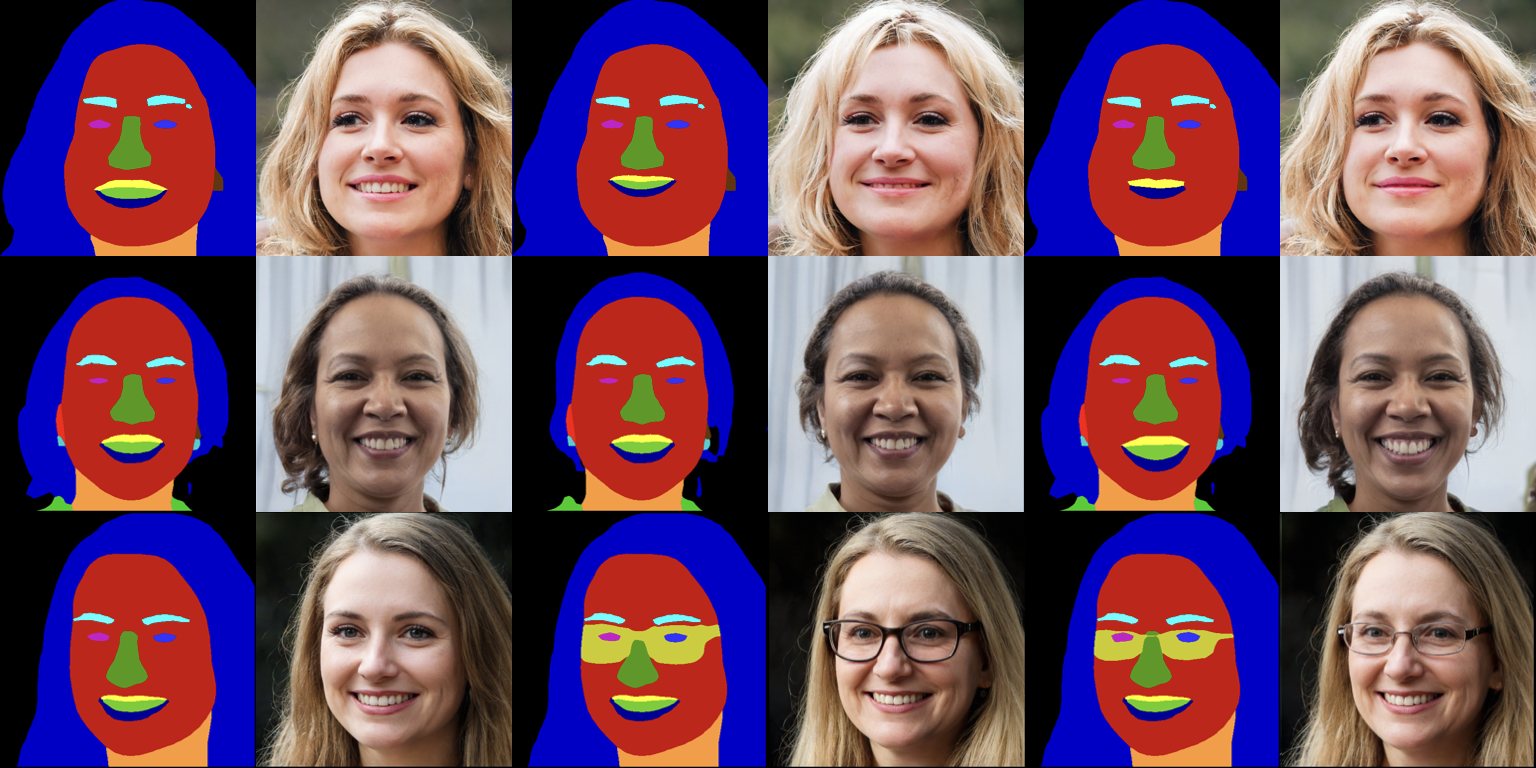}
\end{minipage}
\label{fig:exp:fixedStyle}
}
\hspace{.01\textwidth}
\subfigure[Fixed semantics]{
\begin{minipage}[t][0.20\textwidth][t]{0.35\textwidth}
\centering
\includegraphics[width=\textwidth]{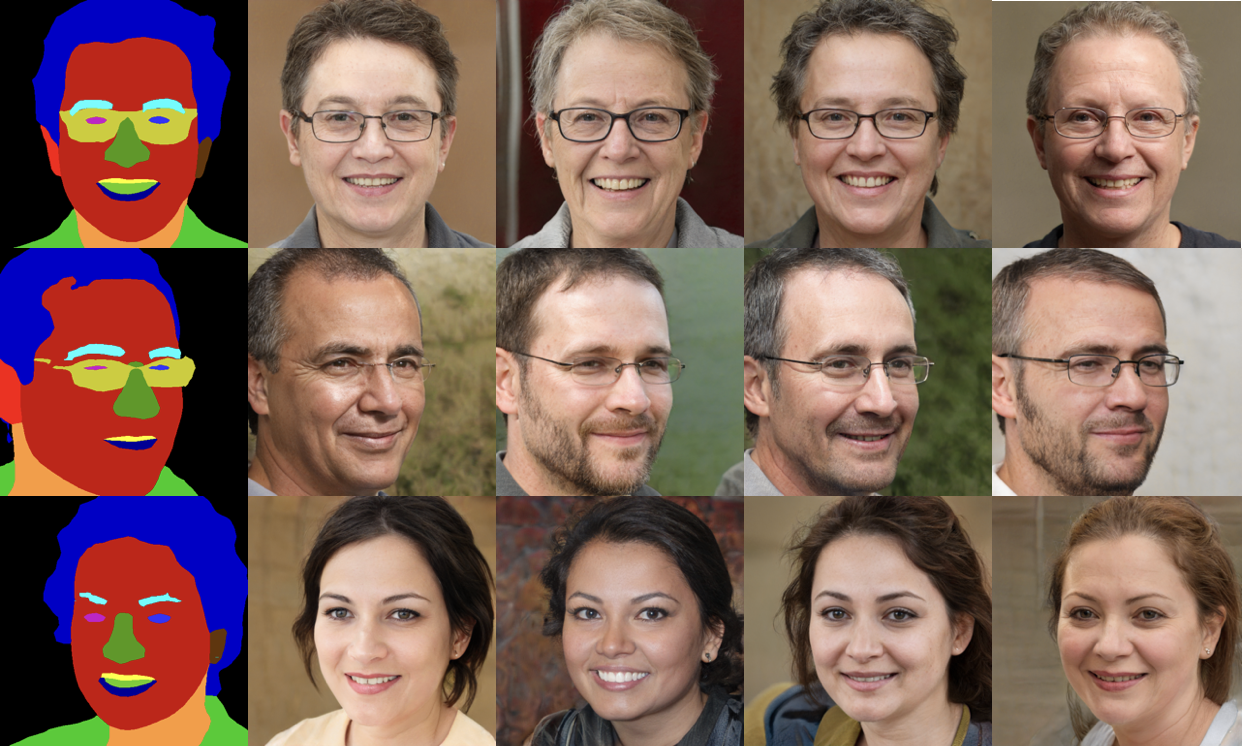}
\label{fig:exp:fixedSemantic}
\end{minipage}
}
\caption{\textbf{Disentanglement of style and semantics.} \emph{(a)} \textbf{Semantic diversity.} \papername{} is able to generate a face with varying smiles, hairstyles, and glasses while preserving the style and identity. \emph{(b)} \textbf{Style diversity.} \papername{} generates a diverse set of styles while only minimally deviating from the input semantic mask.}
\label{fig:exp:disentanglement}
\vspace{-0.5em}
\end{figure}

\myparagraph{Disentanglement of style and semantics.} 
We show qualitatively that our model learns to disentangle style and semantics when generating or editing an image by keeping one dimension fixed while varying the other.
Given a semantic mask, we randomly sample 4 different style codes from the style latent space of \papername{} and generate their corresponding images (\cref{fig:exp:fixedSemantic}), which are diverse in style but consistent with the input semantic mask. Similarly, given an image, we obtain its style code and apply different edits to its semantic mask to generate images with the same style but different semantics (\cref{fig:exp:fixedStyle}).
Both results in \cref{fig:exp:disentanglement} show that our model has learned to disentangle the semantic and style codes for an image.  
Though we achieve a high degree of disentanglement, some factors still remain entangled between the semantic and style spaces, such as long hair and gender; this can be explained as we train with only GAN-generated images. Existing GAN-based approaches are known to not cover all modes of the underlying data distribution \citep{Pei2021Alleviating,Goodfellow2016Nips,Liu2020Diverse}. Perfect disentanglement, even with real images, remains challenging due to the inherent bias of the datasets \citep{Kortylewski2019Analyzing} and the sample complexity \citep{Locatello2019Challenging}.

\myparagraph{Semantic editing.} 
We evaluate the semantic editing capability of our model qualitatively against MaskGAN~\citep{Lee2020MaskGAN} and SPADE~\citep{Park2019Semantic}. We also compare our method against StyleFlow~\citep{Abdal2021StyleFlow} to show that attribute-based methods are unable to apply controlled semantic editing. 
We do not compare our model against unconditional and text-based models since the former lack interactive editing and the latter require very targeted text for highly controlled semantic editing. \cref{fig:exp:semantic_editing_qualatative} clearly shows that our model is much better in terms of visual quality and identity preservation compared to previous semantic editing methods (MaskGAN~\citep{Lee2020MaskGAN}, SPADE~\citep{Park2019Semantic}). \cref{fig:exp:semantic_editing_qualatative} also shows that attribute-based methods (StyleFlow~\citep{Abdal2021StyleFlow}) allow for some semantic edits like smile and glasses, but without controllability over their shape.

\begin{figure}[t]
\centering
\subfigure[Joint style and semantic editing]{
\begin{minipage}[t][0.20\textwidth][t]{0.31\textwidth}
\centering
\includegraphics[width=\textwidth]{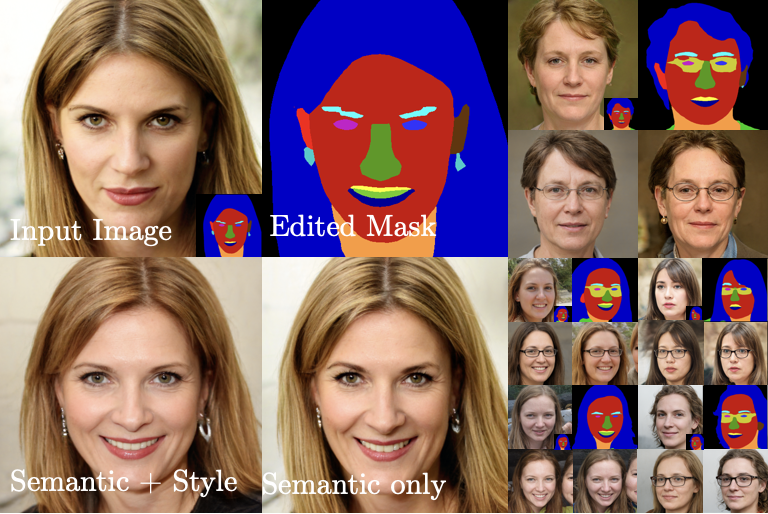}
\end{minipage}
\label{fig:exp:structure_and_style:left}
}
\hspace{0.5pt}
\subfigure[Style editing]{
\begin{minipage}[t][0.20\textwidth][t]{0.62\textwidth}
\centering
\includegraphics[width=\textwidth]{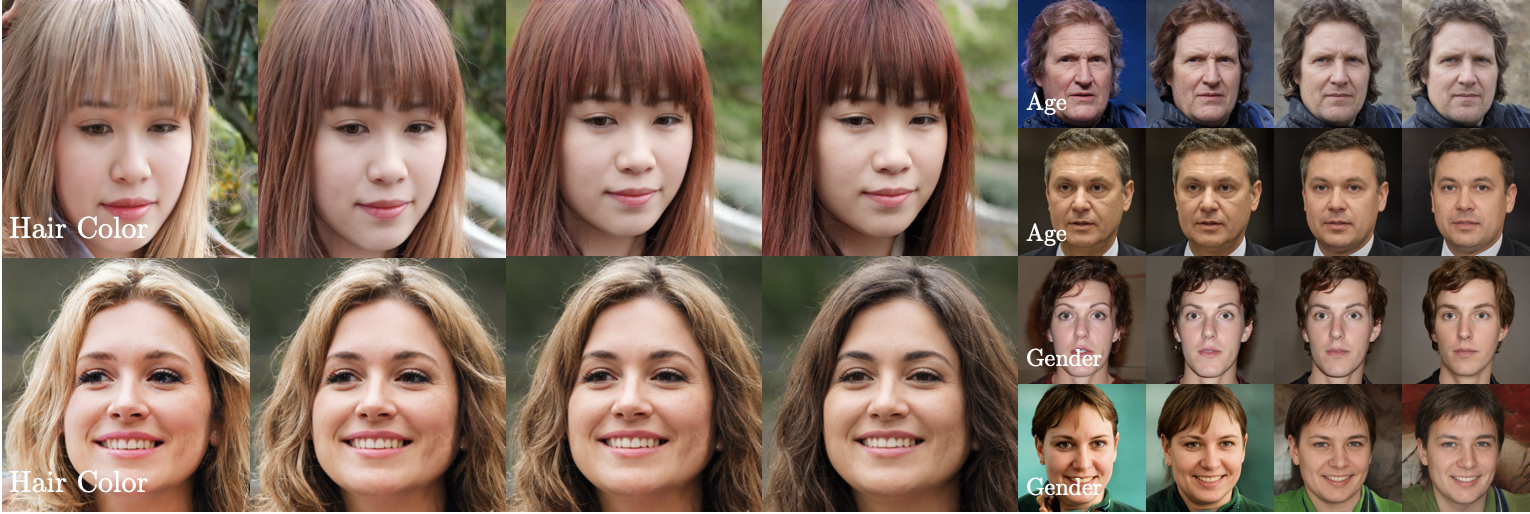}
\label{fig:exp:structure_and_style:right}
\end{minipage}
}
\caption{\emph{(a)} \textbf{Semantic and style editing.} \papername{} enables diverse semantic and style edits like adding a smile (sem) \& changing hair color (sty), adding glasses (sem) \& changing gender (sty), \etc %
\emph{(b)} \textbf{Style editing.} \papername{} is able to modify style attributes like hair color, age, and gender while preserving the identity and being faithful to the input semantics.}
\label{fig:exp:structure_and_style}
\vspace{-0.5em}
\end{figure}

\myparagraph{Style editing.}
\cref{fig:exp:structure_and_style:right} shows the results of \papername{} for different style edits by interpolating in the style latent space. Our model can apply fine-grained style edits like changing hair color and more general edits like changing gender and age while staying truthful to the semantic mask. Since our style space itself is not disentangled by design, multiple attributes can change during interpolation. Doing similar edits with other semantic methods via style transfer requires manually searching for a target image differing in only one attribute. 

\myparagraph{Semantic and style editing.}
We highlight the flexibility of our model in performing edits in multiple spaces, \ie style \emph{and} semantic space. 
\cref{fig:exp:structure_and_style:left} shows that even when we edit multiple attributes in the semantic space and then perform an edit in the style space, our model is still able to preserve the identity and has excellent visual fidelity. Our work is one of the first to allow for joint editing of style and semantics with a high level of control, compared to attribute-based methods that only allow for controlled style edits (\cref{fig:related_work:control_exp:left}) and semantic-based methods that only for controlled semantic edits (\cref{fig:related_work:control_exp:right}).

\begin{figure}[t]
\centering
\subfigure[Semantic editing on real images]{
\begin{minipage}[t][0.20\textwidth][t]{0.30\textwidth}
\centering
\includegraphics[width=0.9\textwidth, height=0.6\textwidth]{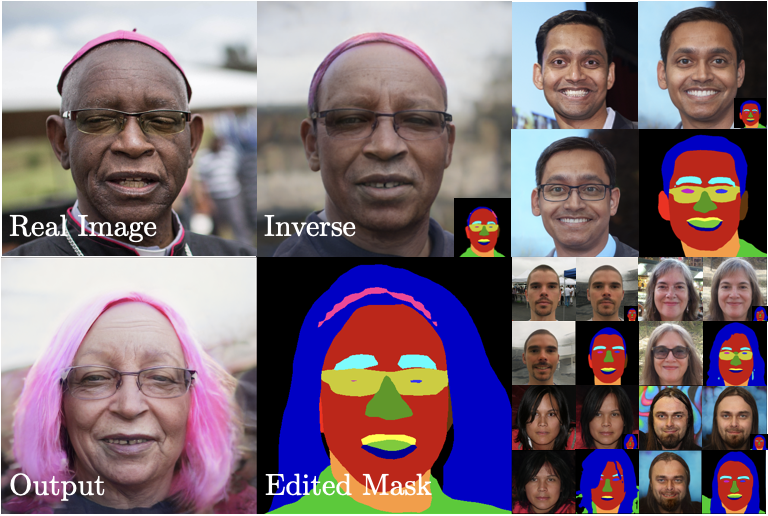}
\end{minipage}
\label{fig:exp:real_result:left}
}
\subfigure[Style editing on real images]{
\begin{minipage}[t][0.20\textwidth][t]{0.30\textwidth}
\centering
\includegraphics[width=0.8\textwidth, height=0.6\textwidth]{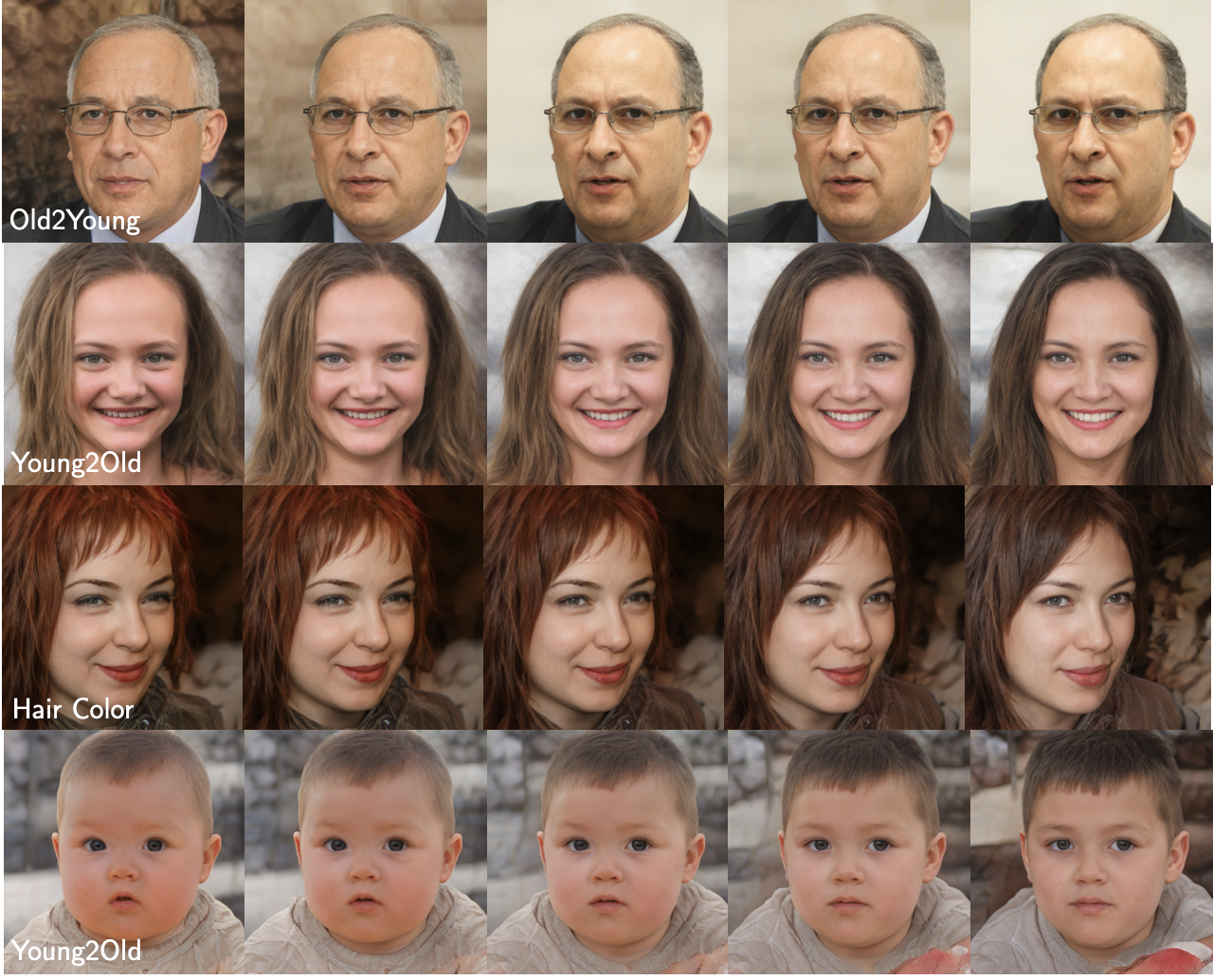}
\end{minipage}
\label{fig:exp:real_result:right}
}
\subfigure[Sequential semantic editing]{
\begin{minipage}[t][0.209\textwidth][t]{0.30\textwidth}
\centering
\includegraphics[width=\textwidth,height=0.63\textwidth]{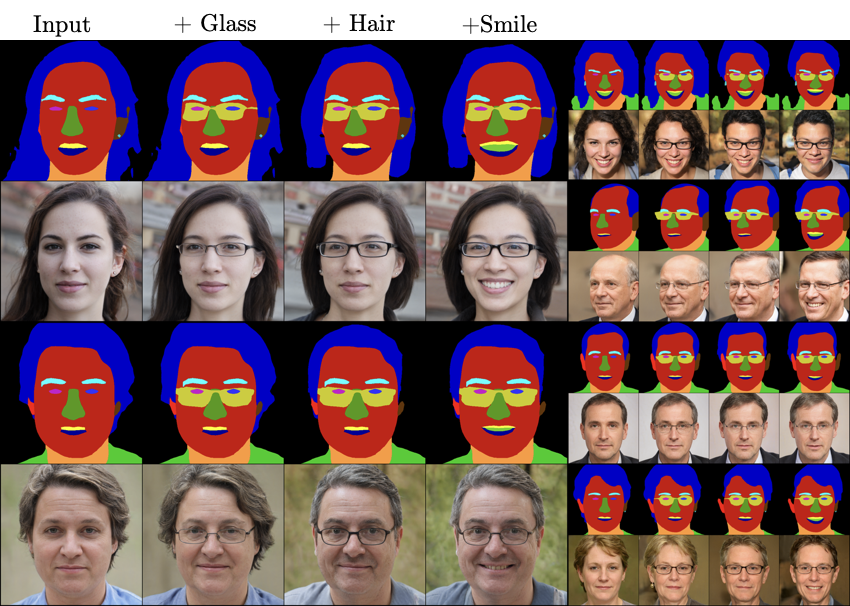}
\end{minipage}
\label{fig:exp:seq_semantic_edit}
}
\caption{\emph{(a)} \textbf{Semantic editing on real images}. \papername{} is able to apply semantic edits like changing hairstyle, adding smile, and adding glasses on real images.
\emph{(b)} \textbf{Style editing on real images.} Our model allows for applying fine style edits like hair color change as well as coarse edits like changing age on real images.
\emph{(c)} \textbf{Sequential semantic editing.} \papername{} is able to perform sequential editing. It outputs latent editing codes, which are amenable for further edits, and the edited images are both identity preserving and realistic.} 
\label{fig:exp:real_result}
\vspace{-0.5em}
\end{figure}

\myparagraph{Sequential semantic editing.}
 We next show the capability of \papername{} on a more extended task that
involves using edited latent vectors from the previous edit to make the next edit. This task is considerably harder since the resultant edited latent vector may not be amenable for further editing \citep{Tov2021Designing}.
\cref{fig:exp:seq_semantic_edit} shows sequential semantic edits like adding glasses, changing the hairstyle, removing glasses, and editing a smile. The results clearly show that \papername{} produces latent codes, which can be edited further. The identity and realism of the input image are preserved even in the case of long-range sequential edits.

\begin{figure}[t]
\centering
\includegraphics[width=\textwidth]{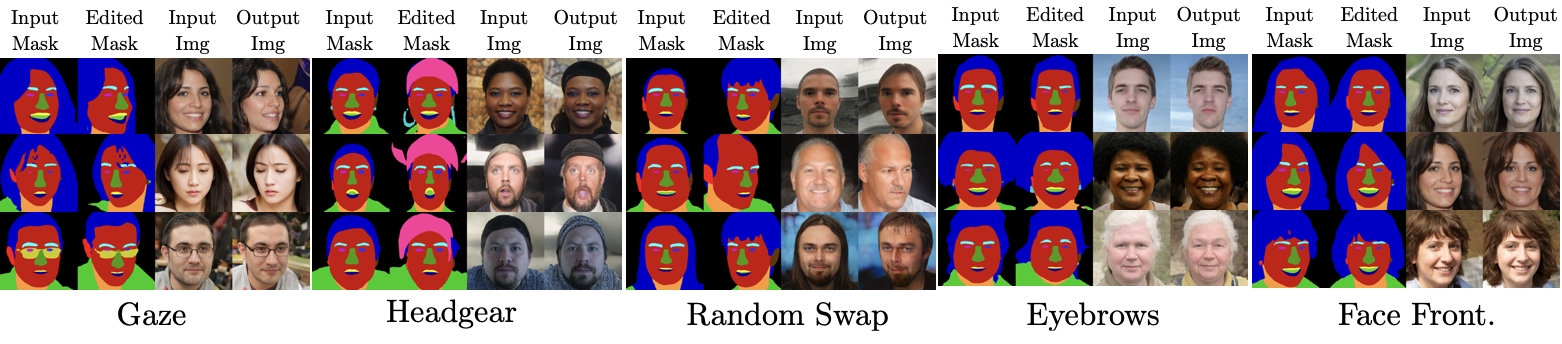}
\vspace{-2em} %
\caption{\textbf{Diverse semantic editing.} \papername{} is capable of a wide variety of semantic edits like changing gaze, changing eyebrows, adding headgear, face frontalization, and random swapping of semantic masks.}
\label{fig:exp:diverse_semantic_editing}
\vspace{-0.5em}
\end{figure}

\myparagraph{Diverse semantic edits.} \cref{fig:exp:diverse_semantic_editing} shows the diverse semantic editing capabilities of \papername{} for tasks like face frontalization, gaze change, \etc{} Even for the difficult case of randomly swapping the masks with another image, which can lead to multiple semantic changes at once, \papername{} preserves the identity of the input while being faithful to the edited mask.

\myparagraph{Real image editing.}
Finally, we show that even though our model is trained on generated images, it can edit real images in both semantic (\cref{fig:exp:real_result:left}) and style spaces (\cref{fig:exp:real_result:right}). We use the e4e~\citep{Tov2021Designing} model to embed the real images into the latent space of StyleGAN2 before editing the images using their corresponding latent vector. 

\myparagraph{Limitations.}
\papername{} is able to disentangle the style and semantics of a given image only to a certain degree. This can mainly be attributed to the fact that present GAN-based methods do not cover all the modes of the underlying data distribution \citep{Pei2021Alleviating,Goodfellow2016Nips,Liu2020Diverse}. 
Even when trained with real images, perfect disentanglement would require an exponential number of samples in the number of factors of variation \citep{Locatello2019Challenging}. 
Also, our model sometimes changes the background while performing style editing, which is mainly due to our style space itself not being disentangled. A simple solution is to use a post-processing optimization step like foreground-background separation or Poisson blending \citep{Perez2003Poisson}. An exciting future direction would be to disentangle the style space itself. This would further increase controllability. 
Moreover, our model only allows for orthogonal changes in the semantic and style attributes. Conflicting semantic and style changes like adding a smile on the mask and then interpolating towards the sad attribute would preserve the semantic mask, only allowing for changes like squinting of eyes, which are orthogonal to semantic changes. How to better handle these conflicting cases remains an open question.  

\section{Conclusions}
We propose $S^2$-Flow, a method to disentangle the latent space of a pretrained generative model to enable facial editing in multiple spaces, namely semantic and style. 
Our novel model design and inductive biases (\emph{semantic} \& \emph{style} consistency) help us to achieve this disentanglement. 
We demonstrate visually and quantitatively that our model outperforms existing semantic-based editing methods 
while also adding controlled style editing capabilities to these models. Further, we illustrate the advantage of semantic editing compared to attribute-based editing. 

\section{Acknowledgments and disclosure of funding}
\label{sec:acknowledgements}
This project has received funding from the
European Research Council (ERC) under the European Union's Horizon 2020 research and innovation programme (grant agreement No.\ 866008).
The project has also been supported in part by the State of Hesse through the cluster projects \enquote{The Third Wave of Artificial Intelligence (3AI)} and \enquote{The Adaptive Mind (TAM)}.

\bibliography{long,egbib}
\newpage
\begin{appendix}
\setlength{\arrayrulewidth}{0.2mm}
\setlength{\tabcolsep}{8pt}
\renewcommand{\arraystretch}{1}
\section{Training Details}\label{appendix:training_details}
\myparagraph{Loss function.} We describe each part of our loss function  (\cref{eqn:full_loss} in the main paper) in detail.
The negative log likelihood loss, $\mathcal{L}_{\text{nll}}$, encourages the network to learn the conditional data distribution of images over semantic masks. It is formally given by
    \begin{equation}
         \mathcal{L}_\text{nll}(\ow) = \log p(\ow|\omask) \ .
    \end{equation}
The semantic consistency loss, $\mathcal{L}_{\text{sm}}$, encourages the network to output an edited latent code \ew\, such that the mask of the generated image, $S(G(\ew))$, is equal to the edited mask \emask\ provided by the user. It is formally written as
\begin{equation}
    \mathcal{L}_\text{sm}(\emask,\ S(G(\ew))) =  \text{CrossEntropy}(\emask, \ S(G(\ew))) \ .
\end{equation}
The pixel-wise image loss, $ \mathcal{L}_\text{\text{img}}$, encourages the network to make the edited area appear similar in style to the original image, \eg, an edited hair/nose texture should be equal to the original hair/nose texture. It formally denoted as 
\begin{equation}
    \mathcal{L}_\text{\text{img}} = \|\eimg - \oimg\|_{2} \otimes M \ ,
\end{equation} 
where $\otimes$ is a pixel-wise multiplication with mask $M$, which restricts the computation to the edited areas. The mask $M$ is computed as the XOR product between the original mask \omask\ and the edited mask \emask. We also tried the XNOR (not XOR) operator and found the XOR operator to perform slightly better in our experiments.

The perceptual loss, $\mathcal{L}_\text{percept}$, encourages the generated image \eimg\ =\ G(\ew) and the original image \oimg\ to be perceptually similar. We use the LPIPS \citep{Zhang2018Unreasonable} loss for this, which is formally given by  
\begin{equation}
    L_{\text{percept}} = \| \phi(\hat{I}) - \phi(\mx)\| _{1} \ ,
\end{equation} 
where $\phi$ denotes the ImageNet\citep{Deng2009Imagenet}-trained VGG \citep{Simonyan2015Very} features.

\myparagraph{Curriculum learning.}  
For training, we rely on a curriculum learning approach where the loss function and the difficulty of the performed edits are increased gradually. Compared to training with the full loss right away, this stabilizes the convergence of our model. 
Concretely, the first 30 epochs are trained only using the negative log-likelihood loss, making the network learn the conditional data distribution. 
For the remaining epochs, the full loss in~\cref{eqn:full_loss} is used; all components of the loss function are equally weighted (\ie~$\lambda_i=1$). 

\begin{wraptable}[10]{r}{6cm}
    \scriptsize
    \vspace{-1.25em}
    \begin{tabularx}{\linewidth}{@{}Xc@{\hspace{0.3cm}}c@{}}
        \toprule
        & & \multirow{2}{*}{\shortstack{Landmark \\distance $\mathcal{D}_\text{L}$~\cite{King2009Dlib}}}\\
        Category & mAcc & \\
        \midrule
        easy &  $\text{mAcc} \geq 0.8$  & $\mathcal{D}_\text{L} \leq 100$ \\
        medium & $0.7 < \text{mAcc} < 0.8$ & $100 <\mathcal{D}_\text{L} < 150$\\
        hard &  $\text{mAcc} \leq 0.7$  & $\mathcal{D}_\text{L} \geq 150$ \\
        \bottomrule
    \end{tabularx}
    \vspace{-1em}
    \caption{Difficulty classification criteria for segmentation edits during training used for curriculum training. See text for details.}
    \label{tbl:exp:category}
\end{wraptable} 

Semantic edits are simulated during training by swapping the ground-truth semantic mask with another one from the training dataset. 
For a given sample and its (source) semantic mask, we categorize the difficulty of swapping it with every other (target) semantic mask in the dataset. 
We categorize the difficulty as a combination of two criteria, as specified in \cref{tbl:exp:category}. 
First, we measure the pixel-wise accuracy between two semantic masks (mAcc).
Second, we compute the landmark distance using \citep{King2009Dlib}, which is the distance between the facial landmarks in the corresponding images. 
Both these criteria help us identify which (target) semantic masks are well aligned with the given (source) sample's semantic mask. Swapping with a well-aligned (target) semantic mask creates a non-noisy semantic mask.  After the initial 30 epochs, we simulate editing during training by swapping the (source) semantic mask with a (target) semantic mask from the easy category. This is done for a duration of 30 epochs. After this, we start swapping with a semantic mask from the medium category for the next 20 epochs. For the remaining 20 epochs, we swap with a semantic mask from the hard category. This strategy helps to stabilize the training process.  

\section{Architecture Details and Ablation}\label{appendix:arch}
Here, we give additional details and ablation experiments, complementing the results shown in the main paper.
Specifically, in~\cref{appendix:arch} we explain the architectural details of our Embedder and CNF blocks used in the proposed network architecture (\cref{fig:method:network_arch}). 
In \cref{appendix:ablations} we give full ablation results on the importance of each component of our loss functions. 

\subsection{Network architecture}\label{appendix:arch-1}
\myparagraph{Embedder.} The embedder is composed of 3 blocks. Each block consists of a ConvolutionBatchNormalization layer followed by a 2D Max-Pool Layer. 
This is followed by an MLP layer, which outputs a vector of size $19\times1$. 
We use ReLU activations after each CNN-BatchNormalization block. 

\myparagraph{CNF block.} 
Each CNF block is made of gate-bias modulation functions called ConcatSquash functions~\citep{Grathwohl2019FFJORD}. 
The gate-bias modulation block consists of 3 linear layers, which modulate the output of the CNF blocks both on the input and the conditioning variable. 
The CNF works by solving an ODE through time, hence the CNF blocks receive time as an input. 
For adding the conditional input to the CNF blocks, we follow~\citep{Abdal2021StyleFlow} and broadcast the time dimension so that it is of the same size as the conditional input. 
The broadcasted time variable and conditional input are then concatenated channel-wise before being fed to the CNF blocks. 

\subsection{Loss ablation}\label{appendix:ablations}
\begin{table}[t]
    \scriptsize
    \begin{tabularx}{\linewidth}{@{}l*5{>{\centering\arraybackslash}X}@{}}
        \toprule
        & \multicolumn{2}{c}{Perceptual Quality} & \multicolumn{2}{c}{Semantic}&Identity\\
        \cmidrule(l{4pt}r{4pt}){2-3} \cmidrule(l{4pt}r{4pt}){4-5} \cmidrule(l{4pt}r{0pt}){6-6}    
        Method  &FID  $(\downarrow)$ & LPIPS $(\downarrow)$ & mIoU $(\uparrow)$ & mAcc $(\uparrow)$ &  ID $(\downarrow)$\ \\
        \midrule
            $\mathcal{L}_{\text{nll}}$ & \sbest 26.50 & 0.17 & \sbest 0.80  & 0.92 & 0.17 \\
            $\mathcal{L}_{\text{nll}} + \mathcal{L}_{\text{sm}}$ & \best\textbf{26.47} & 0.18 & \best \textbf{0.84}  & \sbest 0.93 & 0.19 \\
            $\mathcal{L}_{\text{nll}}+ \mathcal{L}_{\text{sm}}+ \mathcal{L}_{\text{img}}$  & 26.60 &  \sbest 0.16 & 0.79 &  0.91 & \sbest 0.16 \\ 
            $\mathcal{L}_{\text{nll}}+\mathcal{L}_{\text{sm}}+ \mathcal{L}_{\text{img}} + \mathcal{L}_{\text{percept}}$ \emph{(ours)} & 26.75 & \best \textbf{0.12} &  0.78 & \best \textbf{0.94} & \best \textbf{0.10}\\
        \bottomrule
        \end{tabularx}
    \vspace{-1em}
    \caption{Impact of different losses on performance metrics for general edit benchmark.}
    \vspace{-0.5em}
    \label{tbl:supp:ablation_losses}
\end{table}

From \cref{tbl:supp:ablation_losses} we can observe that the  perceptual loss function ($\mathcal{L}_\text{percept}$) is most helpful in terms of identity preservation and the LPIPS \citep{Zhang2018Unreasonable}.
$\mathcal{L}_\text{sm}$ improves the mIoU and mAcc scores compared to $\mathcal{L}_\text{nll}$, indicating that the model learns to be more faithful to the edited semantic mask. 
Surprisingly, we find  $\mathcal{L}_\text{img}$ helps only marginally over $\mathcal{L}_\text{nll} + \mathcal{L}_\text{sm}$ in terms of the LPIPS and ID preservation. A possible reason for this can be the use of masking the $\mathcal{L}_{\text{img}}$ loss to only the edited regions. Our final loss function is much better on three key metrics of LPIPS, ID, and mAcc, while being slightly worse in terms of mIoU and FID scores compared to others.

\begin{figure}[t]
        \includegraphics[width=\textwidth]{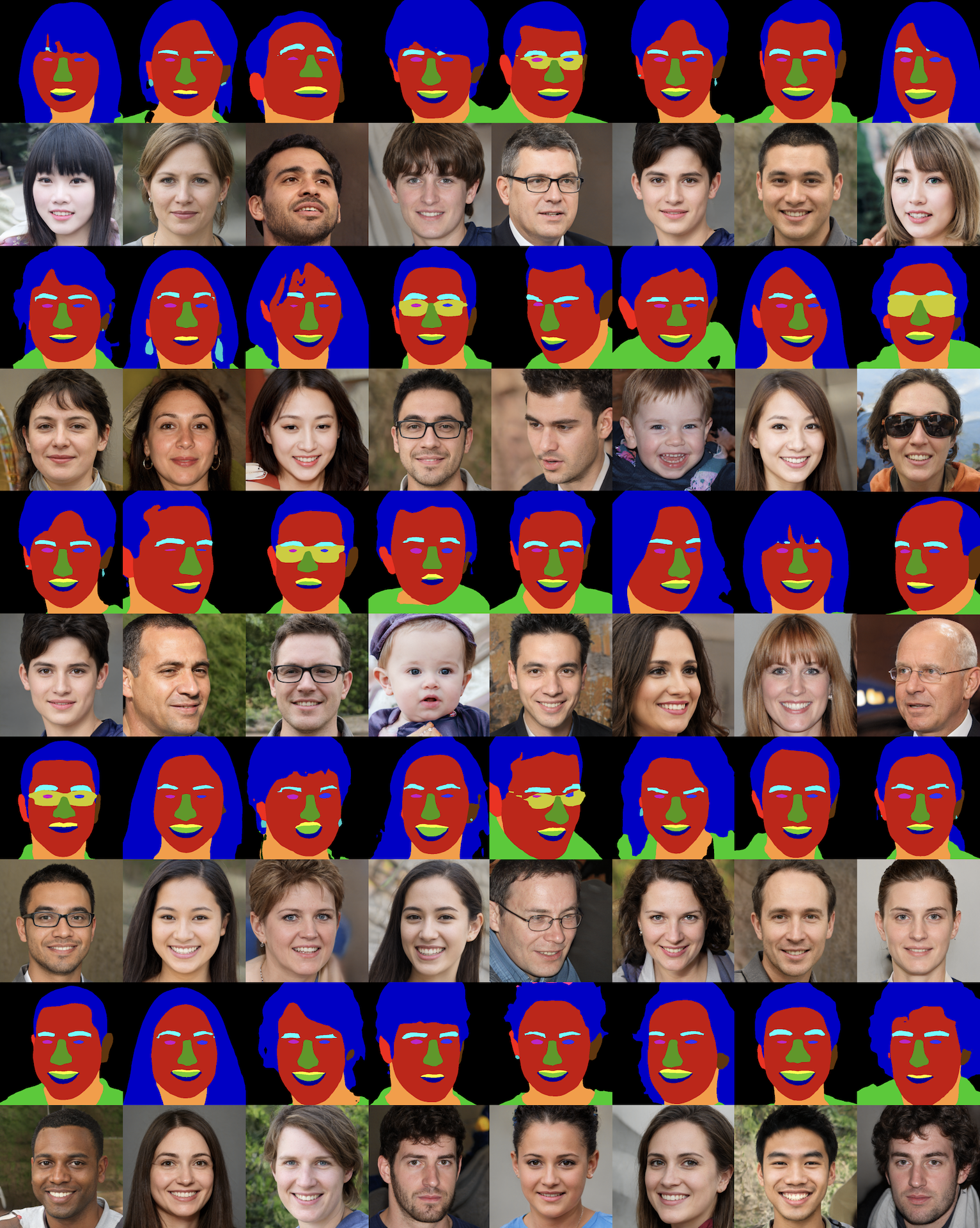} 
        \vspace{-1.5em}
        \caption{\textbf{Conditional generation.} Given a semantic mask \omask, \papername{} can generate highly realistic-looking images, which are pixel precise to the input mask \omask. In comparison, this high level of controllability is not possible with attribute-based, \eg StyleFlow \citep{Abdal2021StyleFlow}, and text-based methods, \eg \citep{Xia2021Tedigan, Li2020Manigan} (see also \cref{fig:related_work:control_exp:left}).}
\label{fig:appendix:cond_gen}
\end{figure}

\begin{figure}[t]
\centering
\subfigure[Fixed Mask]{
\begin{minipage}[t][0.19\textwidth][t]{0.42\textwidth}
\centering
\includegraphics[width=\textwidth]{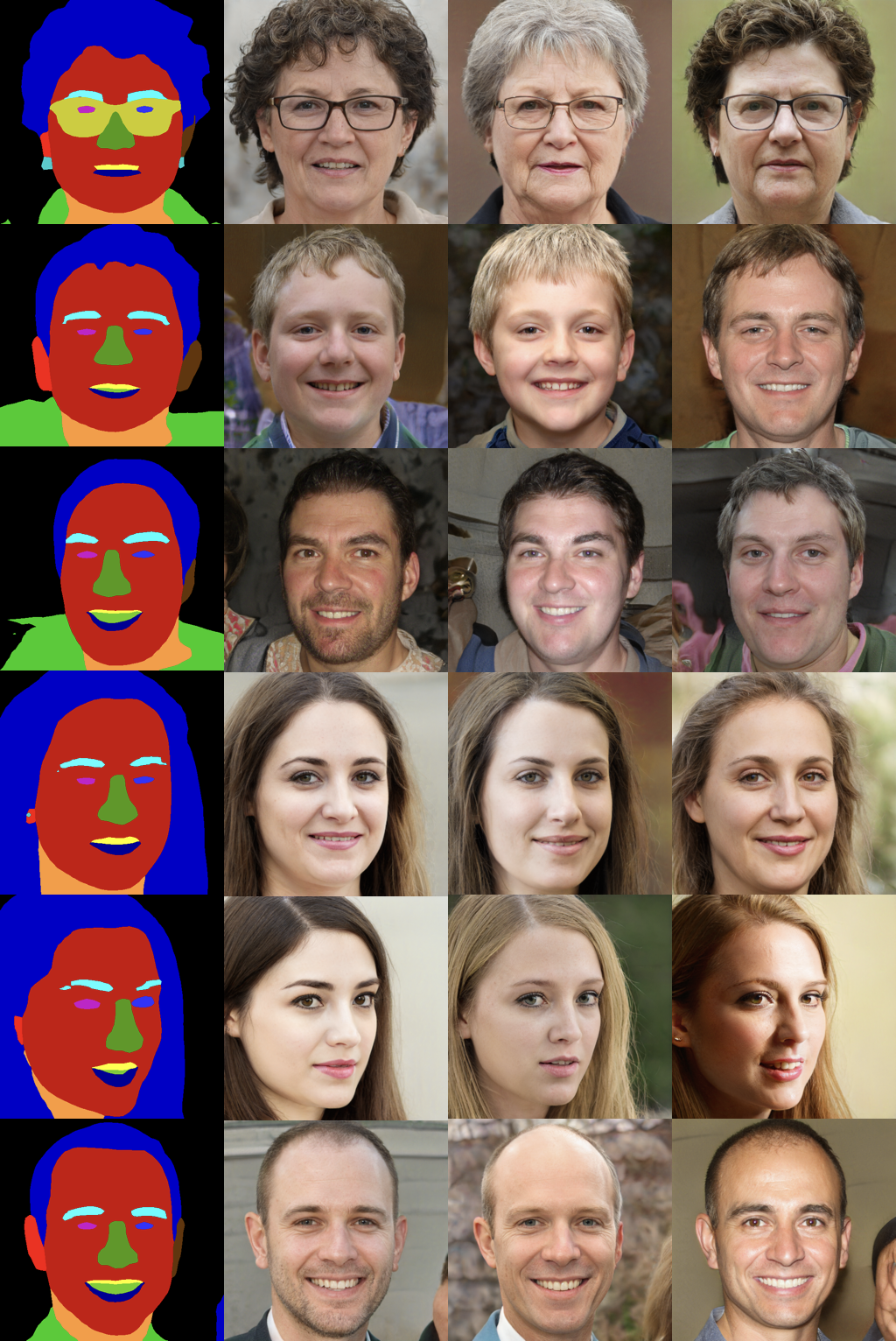}
\label{fig:appendix:fixedSemantic}
\end{minipage}
}
\hspace{10pt}
\subfigure[Fixed Style]{
\begin{minipage}[t][0.19\textwidth][t]{0.42\textwidth}
\centering
\includegraphics[width=\textwidth]{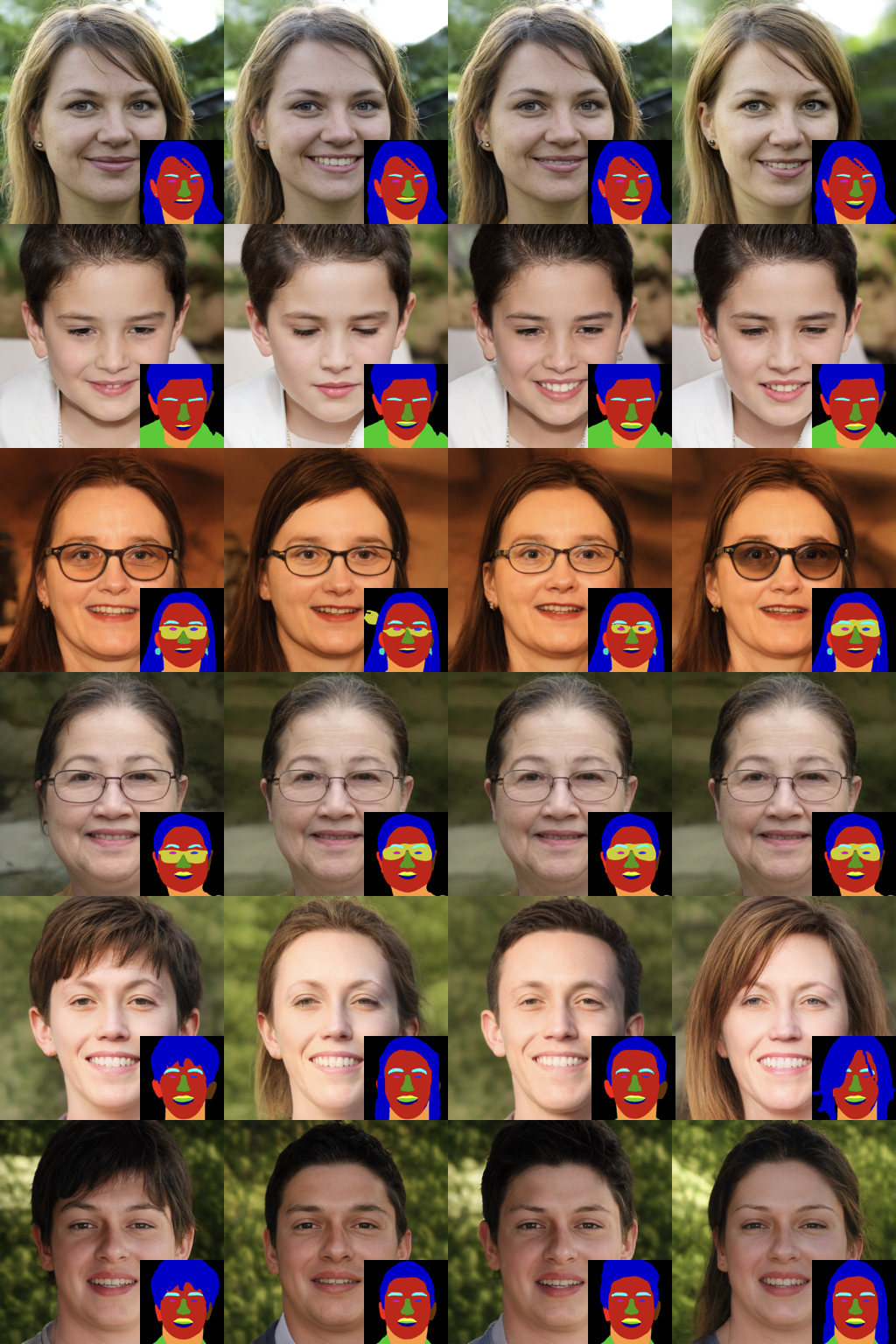}
\label{fig:appendix:fixedStyle}
\end{minipage}
}
\caption{\textbf{Disentanglement of semantics and style.} \emph{(a)} \textbf{Style diversity.} \papername{} generates a diverse set of styles while only minimally deviating from the semantic mask. \emph{(row [1-2])} shows age variations with the same semantic mask and  \emph{(row [3-5])} shows hair color variations while keeping the semantic mask fixed. \emph{(b)} \textbf{Semantic diversity.} \papername{} is able to generate a face with varying smiles \emph{(row [1-2])}, glasses \emph{(row [3-4])}, and hairstyle \emph{(row [5-6])} while keeping the style fixed and preserving identity.}
\label{fig:appendix:disentanglement}
\vspace{-1em}
\end{figure}

\begin{figure}[t]
    \centering
        \includegraphics[width=0.95\textwidth]{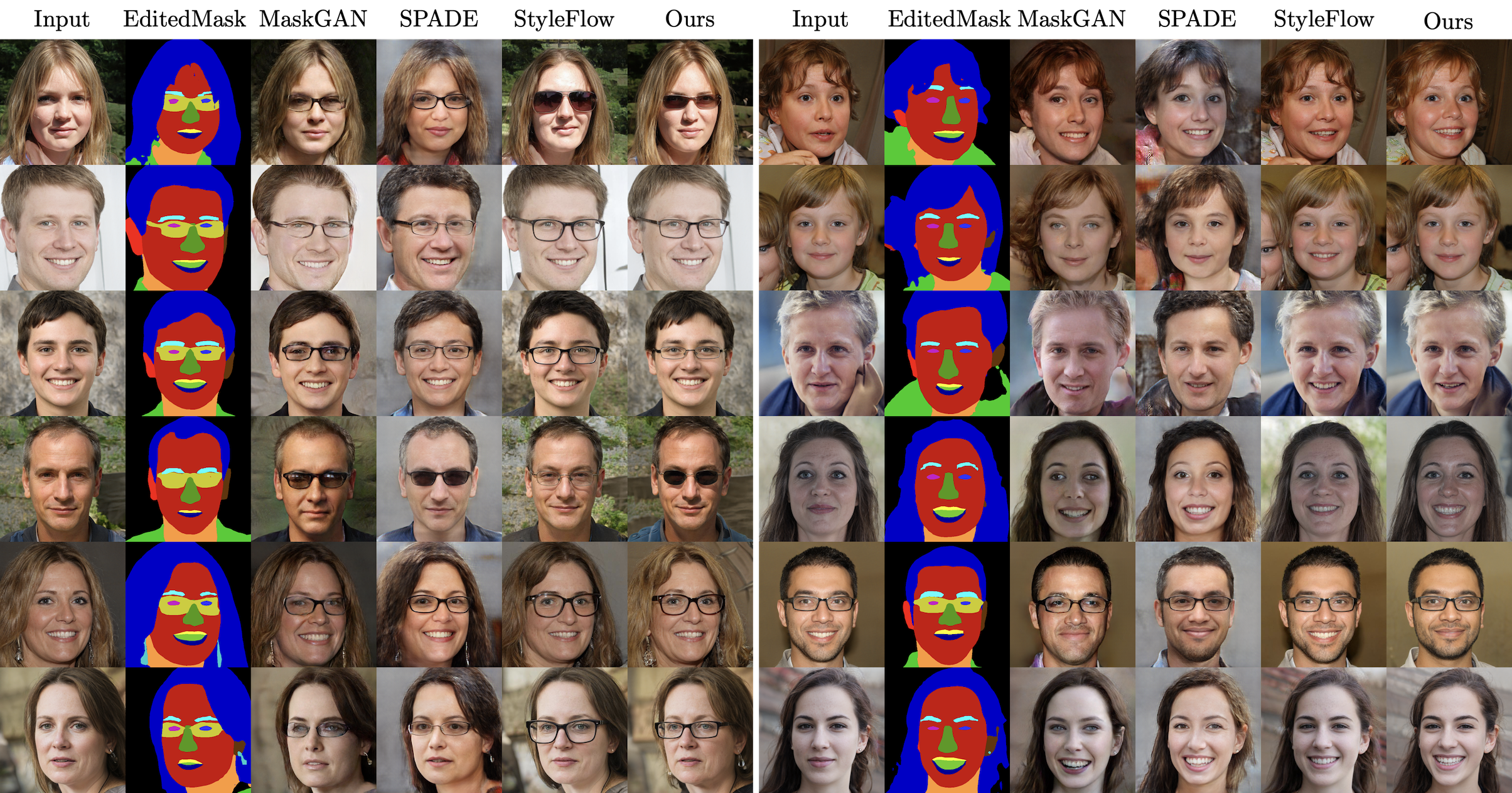} 
    \vspace{-0.5em}
    \caption{\textbf{Semantic editing.} \papername{} is able to apply semantic edits that are more controllable (this is particularly visible for adding glasses) than attributed-based methods, \eg StyleFlow~\citep{Abdal2021StyleFlow}, and has higher identity preservation and realism compared to semantic-based methods, \eg MaskGAN \citep{Lee2020MaskGAN} and SPADE \citep{Park2019Semantic}.}
\vspace{-1.5em}
\label{fig:appendix:semantic_cmp}
\end{figure}

\begin{figure}[t]
    \centering
        \includegraphics[width=\textwidth]{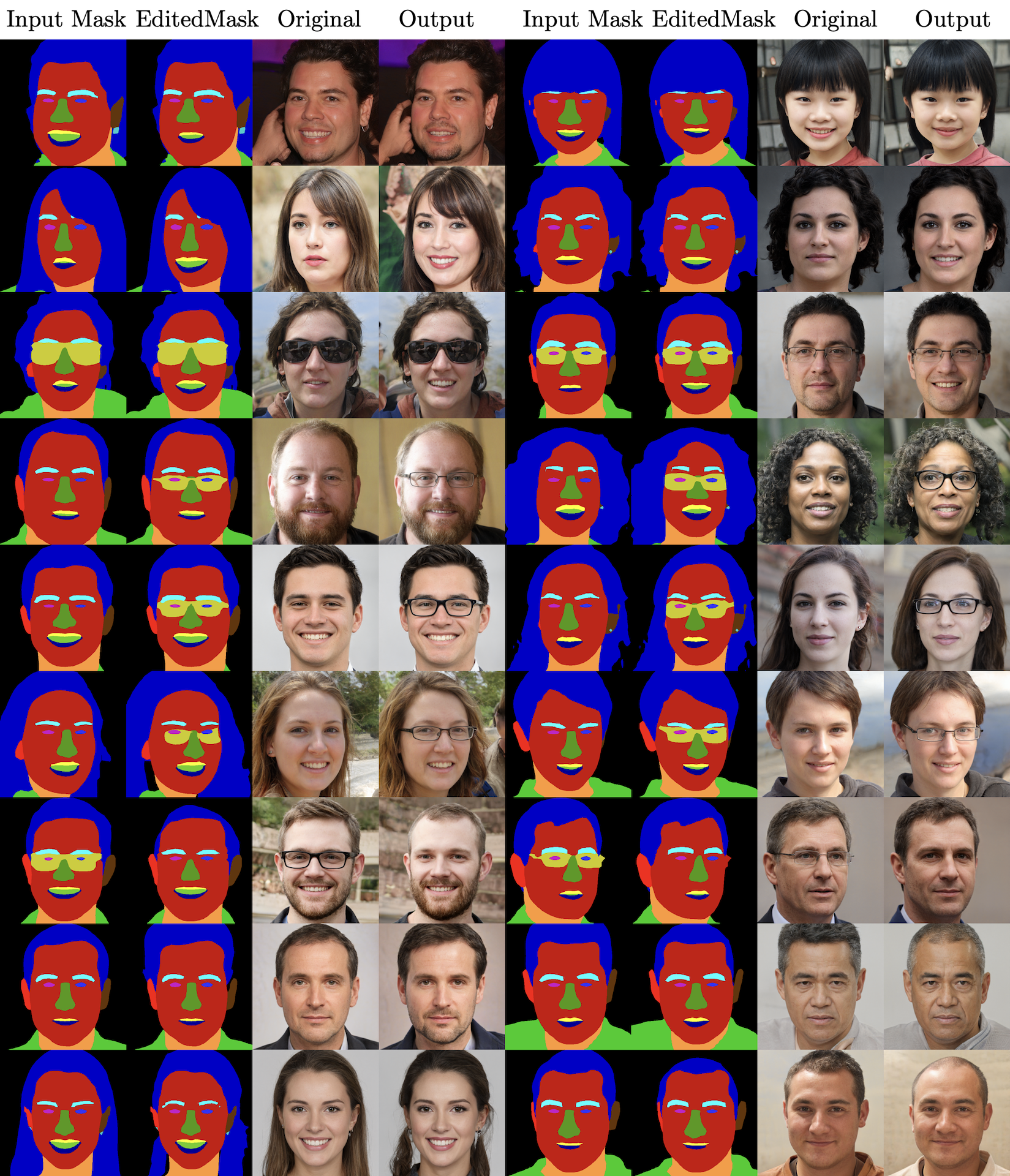} 
    \vspace{-1.5em}
    \caption{\textbf{Semantic editing on generated images.} Our model is able to provide a wide variety of semantic edits like smile change \emph{(row [1-3])}, adding glasses \emph{(row [4-5, 6-left])}, removing glasses \emph{(row [ 6-right, 7])}, and changing hairstyle \emph{(row [8-9])} on GAN-generated images. All edits are identity preserving and have a high visual realism. Even for a noisy edited mask \emph{(row [6, 7]-left)} with multiple changes, \papername{} is able to make high-quality edits that are faithful to the input mask.}
\vspace{-1.5em}
\label{fig:appendix:semantic_fake}
\end{figure}

\begin{figure}[t]
    \centering
        \includegraphics[width=\textwidth]{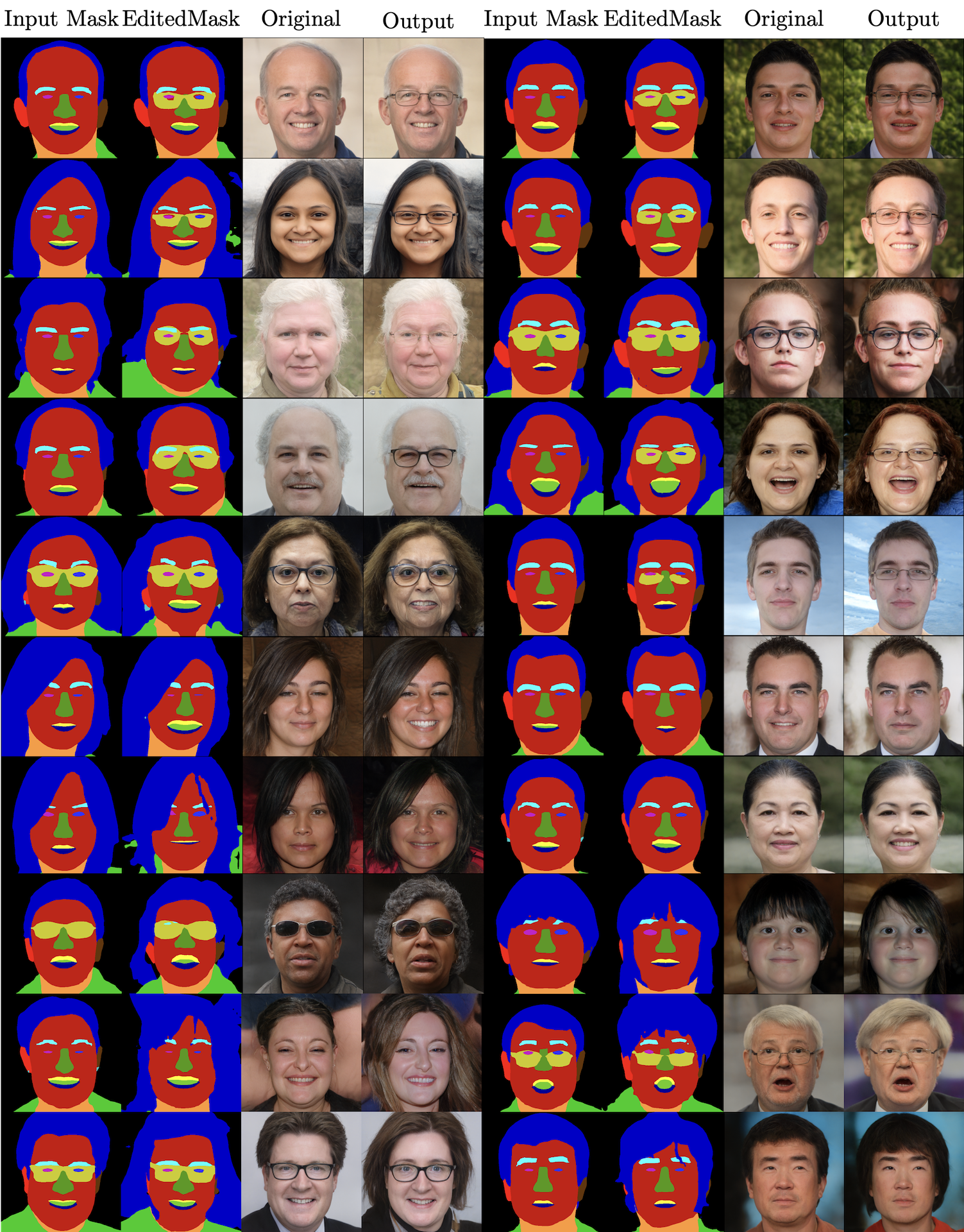} 
    \vspace{-1.5em}
    \caption{\textbf{Semantic editing on real images.} \papername{} can make a wide variety of semantic edits like adding glasses \emph{(row [1-4, 5-right])}, smile editing \emph{(row [5-left, 6-7])}, and hairstyle editing \emph{(row [8-10])} on real images, even when the model is never trained on real images. Even with a noisy edited mask \emph{(row [3-left, 5-right])}, the model is able to preserve identity while only slightly changing some style attributes (background color).}
\vspace{-1.5em}
\label{fig:appendix:semantic_real}
\end{figure}

\section{Additional Qualitative Results}
We provide numerous additional visual results to show the editing capabilities of our method.

In \cref{appendix:cond_gen}, we show visual results for the task of conditional generation. \cref{appendix:disentanglement,appendix:semantic_edit,appendix:style_edit} show additional qualitative results for the task of disentanglement, semantic and style editing.

\subsection{Conditional generation}
\label{appendix:cond_gen}
\cref{fig:appendix:cond_gen}\ shows that \papername{} is able to generate realistic-looking images that are pixel precise to the given input mask. This is in contrast to other attributed-based methods, \eg StyleFlow~\citep{Abdal2021StyleFlow}, and text-based methods, \eg ManiGAN~\citep{Li2020Manigan} or TediGAN~\citep{Xia2021Tedigan}, which provide little control over how the generated images should look (\cref{fig:related_work:control_exp}). Though semantic-based methods like MaskGAN~\citep{Lee2020MaskGAN} and SPADE~\citep{Park2019Semantic} are also able to perform controlled conditional generation. \cref{fig:appendix:semantic_cmp} shows results in addition to \cref{fig:exp:semantic_editing_qualatative} for these methods, which showcase that semantic-based methods lack realism when generating images, which was also seen quantitatively in \cref{tbl:exp:smile_edit_benchmark,tbl:exp:general_benchmark}. 
\subsection{Disentanglement}
\label{appendix:disentanglement}
One of the key ideas of \papername{} is to disentangle the style and semantic spaces to provide controlled edits in both spaces. 
\cref{fig:appendix:disentanglement} shows additional results to \cref{fig:exp:disentanglement}, verifying the disentanglement of \papername{}.
\cref{fig:appendix:fixedSemantic} shows that the \papername{} can generate images with diverse style while being faithful to the semantic mask. 
On the other hand, \cref{fig:appendix:fixedStyle} shows the results for the case when the style code is fixed and the semantic mask is changed. This is equivalent to semantically editing an image. \papername{} is able to generate images that are faithful to the input mask while preserving the identity and style of the input image (see \cref{fig:appendix:fixedStyle}).

\subsection{Semantic editing}
\label{appendix:semantic_edit}
Since \papername{} is primarily a semantic-based editing method, we provide several additional results on the task of semantic editing. 
First, \cref{fig:appendix:semantic_cmp} provides results in addition to \cref{fig:exp:semantic_editing_qualatative} for comparing \papername{} against its peers, namely MaskGAN \citep{Lee2020MaskGAN} and SPADE \citep{Park2019Semantic}. We also include a comparison against an attribute-based method, StyleFlow \citep{Abdal2021StyleFlow}, to further show the superiority and need of semantic-based methods for controlled semantic editing. \cref{fig:appendix:semantic_cmp} shows that \papername{} is superior in terms of visual quality to MaskGAN and SPADE while being more faithful to the edited input mask than StyleFlow, leading to more controlled editing; this was also supported quantitatively in \cref{tbl:exp:general_benchmark,tbl:exp:smile_edit_benchmark}. 

We then show results for semantic editing of fake images (\cref{fig:appendix:semantic_fake}), and additional results to \cref{fig:exp:real_result:left} for editing real images (\cref{fig:appendix:semantic_real}) on various semantic edits like editing a smile, changing glasses, and hairstyle. We want to point out that sometimes when the edited semantic mask is noisy\footnote{Noisy semantic masks happen due to swapping semantic parts between images that are not well aligned.} (\cref{fig:appendix:semantic_fake} row [6,7]-left and \cref{fig:appendix:semantic_real} row [3-left]), it can lead to slight changes in the background or other style attributes. However, the edited image by \papername{} still preserves the identity of the original image to a very high degree. 

\subsection{Style editing}\label{appendix:style_edit}
\begin{figure}[t]
    \centering
    \includegraphics[width=\textwidth]{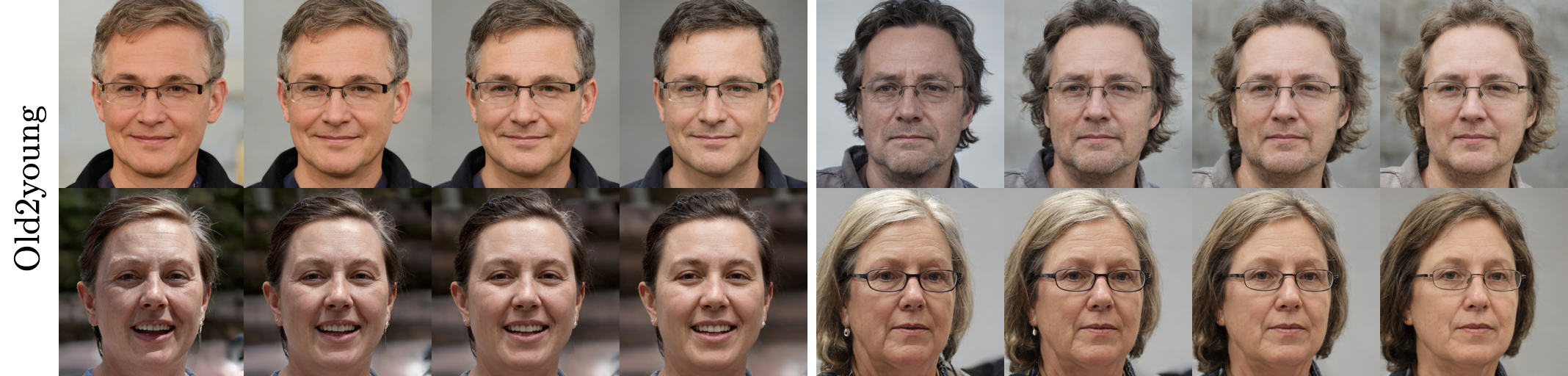}
    \includegraphics[width=\textwidth]{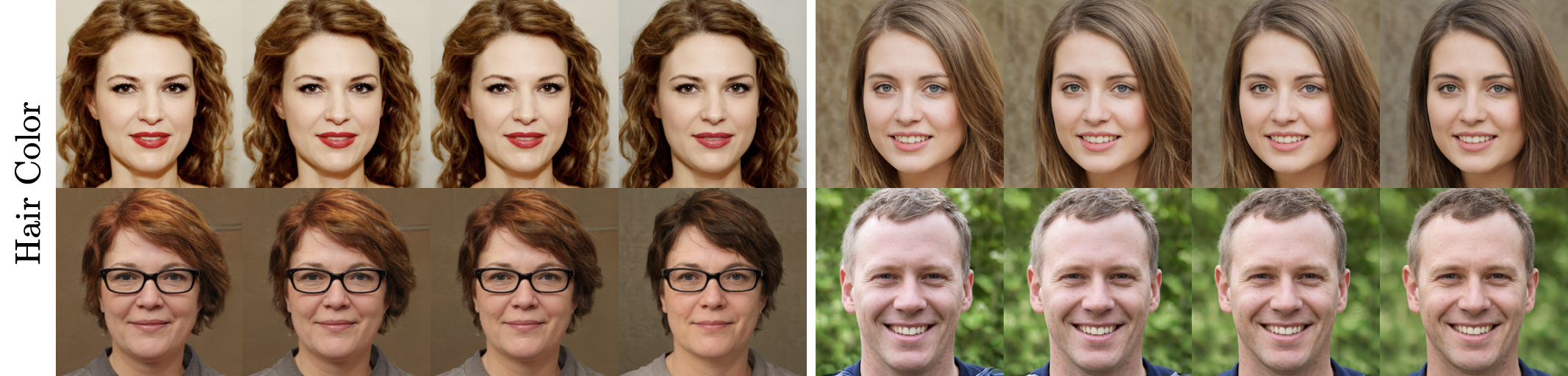}
    \includegraphics[width=\textwidth]{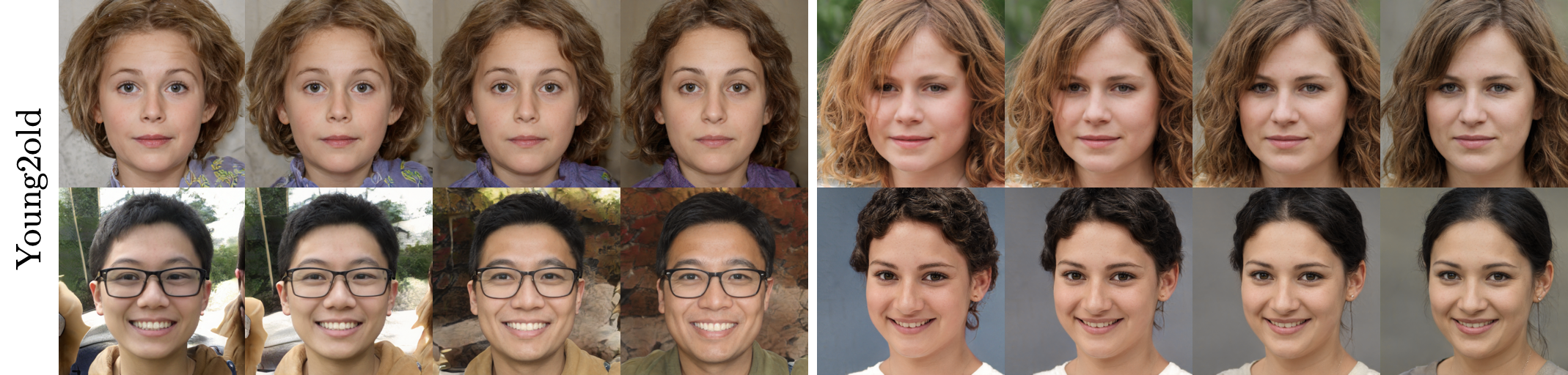}
    \vspace{-1.5em}
    \caption{\textbf{Style editing of generated images.} Our model is able to apply fine style edits like hair color, as well as broad style edits like  age (young2old and
old2young), with high degree of identity preservation and realism. Some erroneous attributes, \eg background, can change when applying style edits because the style space of \papername{} is not disentangled itself.}
    \label{fig:appendix:fake_style_img}
\end{figure}

\begin{figure}[t]
    \centering
    \includegraphics[width=\textwidth]{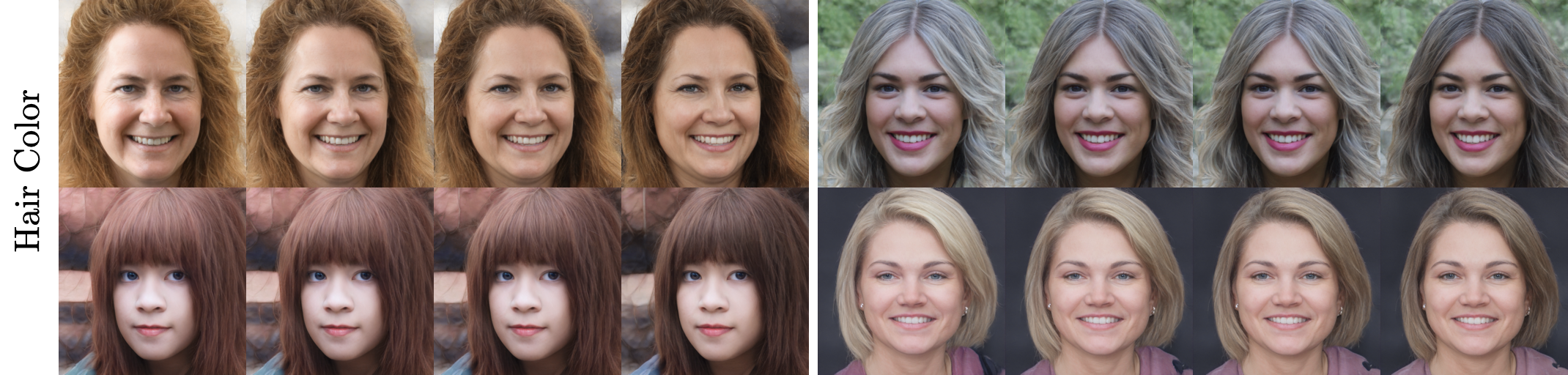}
    \includegraphics[width=\textwidth]{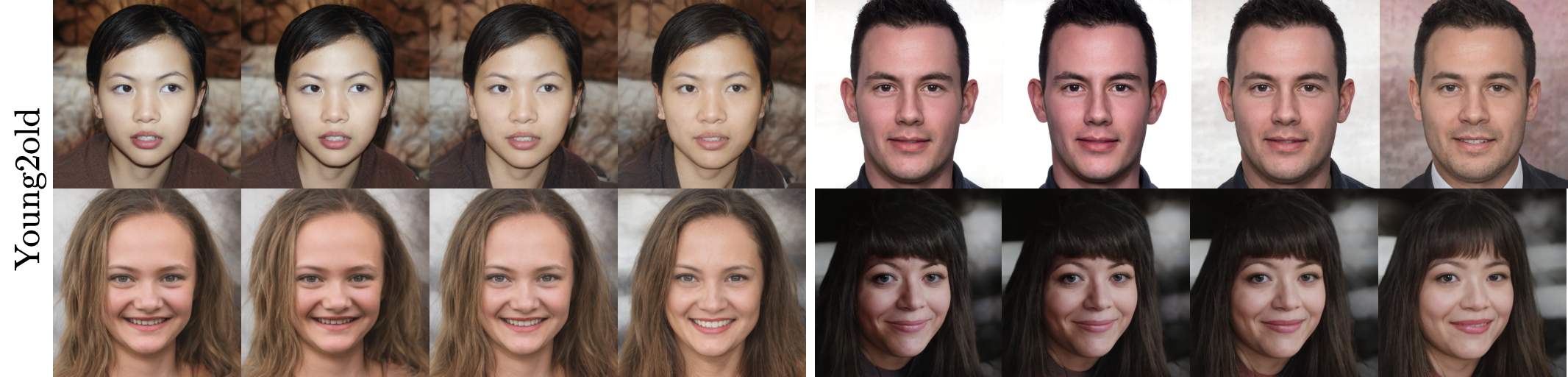}
    \includegraphics[width=\textwidth]{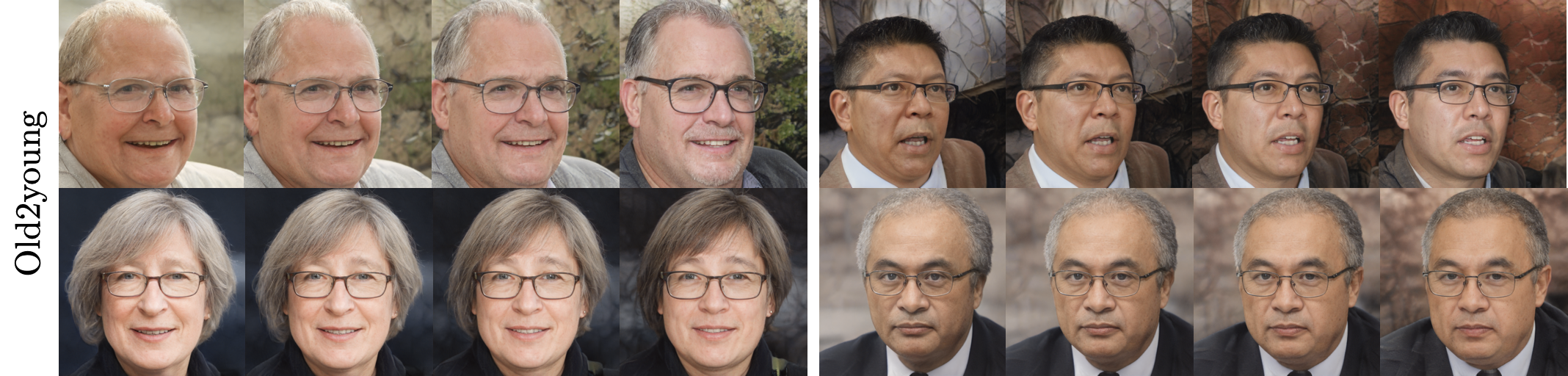}
    \vspace{-1.5em}
    \caption{\textbf{Style editing of real images.} Our model is able to apply fine style edits like changing hair color, as well as broad style edits like age edits (young2old and old2young) on real images with a high degree of identity preservation, even though the model is never trained on real images.}
    \label{fig:appendix:real_style_img}
\end{figure}

\begin{figure}[t]
\centering
\includegraphics[width=0.8\textwidth]{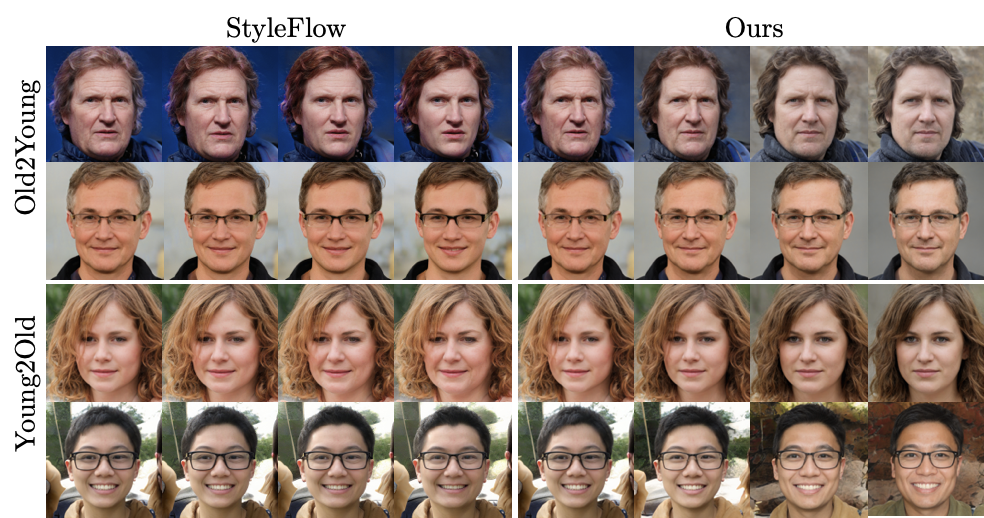}
\caption{
\textbf{Style editing comparison.} Age style editing comparison between StyleFlow and \papername{}. Though \papername{} is never trained explicity with style attributes, it is still is comparable to StyleFlow, a model specifically trained for style editing. }
\label{fig:supp:sty_cmp}
\vspace{-1.5em}
\end{figure}

\cref{fig:appendix:fake_style_img,fig:appendix:real_style_img} show additional results to \cref{fig:exp:structure_and_style:right,fig:exp:real_result:right} for style editing of fake and real images, respectively. 
\papername{} uses simple linear interpolation in its style space for applying style edits. It is able to apply fine edits like changing hair color and broad edits like modifying age. While making style edits, our method preserves the person's identity and only minimally changes the semantics of a given image. 
We want to point out that since \papername{}'s style space is not disentangled, broad style edits like age can cause multiple attributes to change (\cref{fig:appendix:real_style_img} row 5 and \cref{fig:appendix:fake_style_img} row 6). 
We can alleviate this issue by disentangling the style space or using non-linear interpolation methods like \citep{Khrulkov2021Latent}.  
In \cref{fig:supp:sty_cmp}, we show results for a style editing comparison with StyleFlow. Though \papername{} is never explicitly trained on style attributes, it still has comparable results with StyleFlow, a model explicitly trained with style attributes. Moreover, \papername{} allows for additional edits like hair color (\cref{fig:exp:structure_and_style:right}), which are not possible with StyleFlow. The results show that our method performs comparably to models that are designed for style editing.


\section{Limitations}\label{appendix:limitation}
\begin{figure}[t]
\centering
\subfigure[Semantic editing -- hair style change]{
\begin{minipage}[t][0.20\textwidth][t]{0.42\textwidth}
\centering
\includegraphics[width=\textwidth]{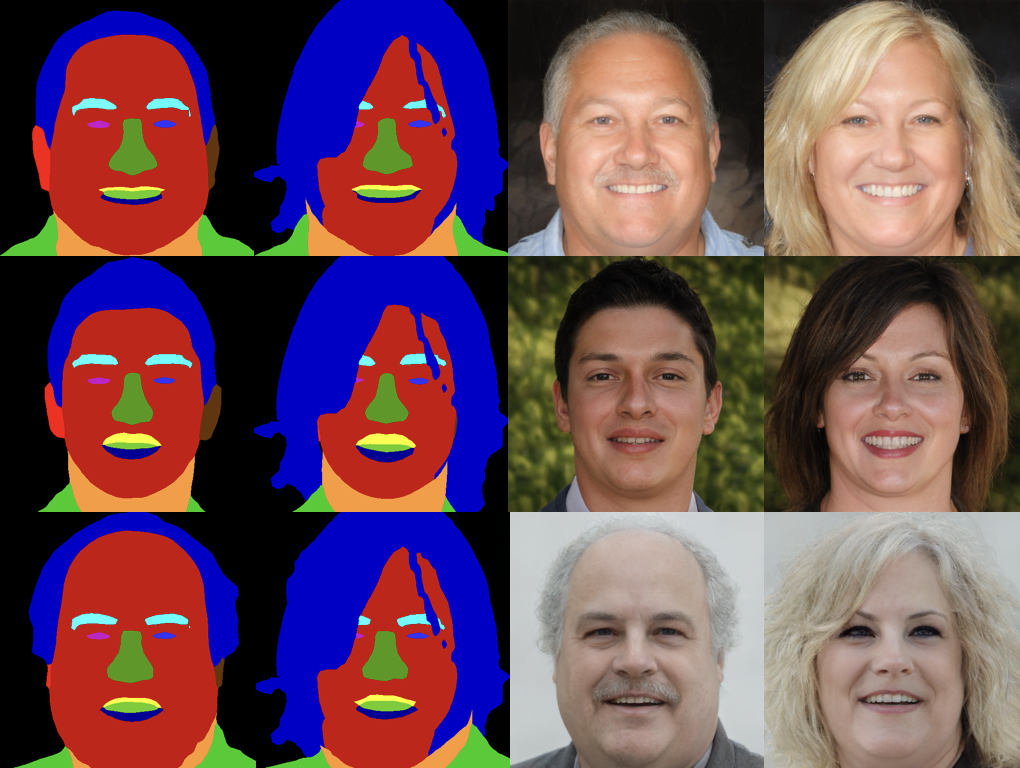}
\label{fig:appendix:sem_sty_entangled}
\end{minipage}
}
\hspace{10pt}
\subfigure[Style editing -- old2young]{
\begin{minipage}[t][0.20\textwidth][t]{0.42\textwidth}
\centering
\includegraphics[width=\textwidth]{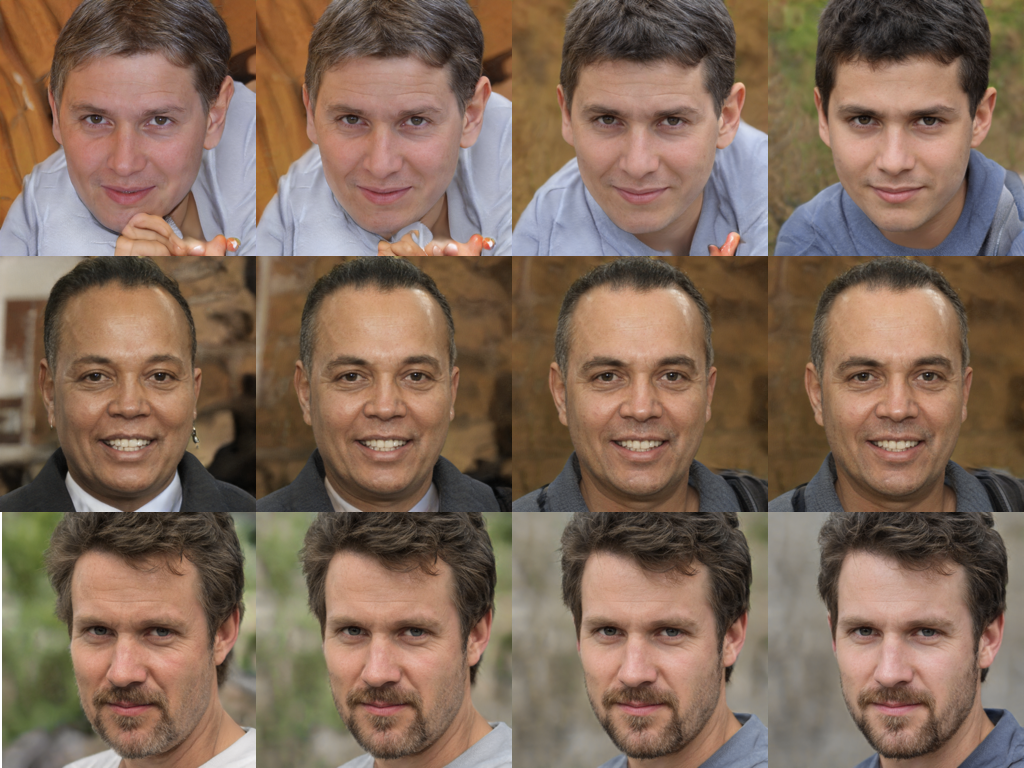}
\label{fig:appendix:sty_entangled}
\end{minipage}
}
\caption{\textbf{Limitations.} \emph{(a)} \textbf{Semantic and style entanglement.} \papername{} is unable to perfectly disentangle long hair (semantic) from attributes like lip and eye makeup (style). This can be attributed to the dataset bias as there exists no or limited samples of men having long hair in the StyleFlow \citep{Abdal2021StyleFlow} dataset with which \papername{} is trained. \emph{(b)} 
\textbf{Style space entanglement.} The style space of \papername{} is entangled, and, when coupled with linear interpolation for performing style edits, it can cause multiple attributes like background, shirt color, \etc to change when applying age editing (old2young).} 
\label{fig:appendix:limitations}
\vspace{-1.5em}
\end{figure}

In this section we provide visual results showing the limitations of \papername{}.
As discussed previously, perfect disentanglement with a limited number of GAN-generated data is impossible \citep{Locatello2019Challenging,Goodfellow2016Nips,Liu2020Diverse,Pei2021Alleviating}. \cref{fig:appendix:sem_sty_entangled} shows that \papername{} is unable to disentangle semantic attributes like long hair with style attributes like eye and lip makeup. This can be attributed to the model being trained on a GAN-generated dataset \citep{Abdal2021StyleFlow}, which contains no or limited examples of men with long hair, making the model associate long hair with the female gender erroneously. Many models suffer from dataset bias, requiring us to be mindful of the dataset with which we train our model. 
The style space of \papername{} is entangled and when coupled with linear interpolation, it causes multiple style attributes to change. \cref{fig:appendix:sty_entangled} shows results where multiple style attributes change when performing age editing (old2young) in the style space of \papername. 
As stated previously, these issues can be resolved by either disentangling the style space of \papername{} or by using non-linear interpolation methods \citep{Khrulkov2021Latent}.

\end{appendix}
\end{document}